\def\year{}\relax

\documentclass[letterpaper]{article} 
\usepackage{aaai}  
\usepackage{times}  
\usepackage{helvet} 
\usepackage{courier}  
\usepackage[hyphens]{url}  
\usepackage{graphicx} 
\usepackage{xcolor}
\usepackage{natbib}  
\usepackage{caption} 
\usepackage{hyperref}
\hypersetup{
    colorlinks=false,
    citebordercolor=green,
    linkbordercolor=red,
    urlbordercolor=cyan,
}

\usepackage[utf8]{inputenc} 
\usepackage{url}            
\usepackage{booktabs}       
\usepackage{amsfonts}       
\usepackage{nicefrac}       
\usepackage{microtype}      
\usepackage{color}
\usepackage{multirow,multicol,booktabs,makecell}
\usepackage{enumitem}
\usepackage{amsmath,amssymb,amsthm,bm}
\usepackage{graphicx}
\usepackage{array}
\usepackage[subrefformat=parens]{subcaption}

\theoremstyle{remark}

\newtheorem{example}{Example}
\theoremstyle{definition}
\newtheorem{definition}{Definition}
\theoremstyle{plain}

\newtheorem{proposition}{Proposition}
\newtheorem{propositionappendix}{Proposition}
\newtheorem{theorem}{Theorem}
\newtheorem{corollary}{Corollary}

\newcommand{\tr}{\mathsf{T}}

\newcommand{\invset}{S}
\newcommand{\stepfunc}{u}

\newcommand{\lyapcond}[2]{{\nabla V(#2)^\tr #1(#2) + \alpha V(#2)}}
\newcommand{\surfcond}[2]{{\nabla C_{\tilde{\invset}}(#2)^\tr #1(#2)}}
\newcommand{\nablanorm}[2]{{\Vert \nabla #1(#2) \Vert_2^2}}

\usepackage{pgfplots}
\pgfplotsset{compat=1.16}
\usepackage{tikz}
\usetikzlibrary{backgrounds}
\usetikzlibrary{arrows}
\usetikzlibrary{shapes,shapes.geometric,shapes.misc}

\pgfdeclarelayer{edgelayer}
\pgfdeclarelayer{nodelayer}
\pgfsetlayers{background,edgelayer,nodelayer,main}

\pgfplotsset{
    /pgfplots/xlabel near ticks/.style={
        /pgfplots/every axis x label/.style={
        at={(ticklabel cs:0.5)},anchor=near ticklabel
        }
    },
    /pgfplots/ylabel near ticks/.style={
        /pgfplots/every axis y label/.style={
        at={(ticklabel cs:0.5)},rotate=90,anchor=near ticklabel}
    }
}

\pgfplotsset{every axis/.append style={
    label style={font=\scriptsize},
    tick label style={font=\tiny},
}}

\pgfplotsset{
    legend image code/.code={
        \draw[mark repeat=2,mark phase=2]
            plot coordinates {
            (0cm,0cm)
            (0.15cm,0cm)        
            (0.3cm,0cm)         
        };%
    }
}

\makeatletter
\newcommand{\mylabeltext}[3][]{%
    \@bsphack%
    \csname phantomsection\endcsname
    \def\tst{#1}%
    \def\labelmarkup{}
    \def\refmarkup{}%
    \ifx\tst\empty\def\@currentlabel{\refmarkup{#2}}{\label{#3}}%
    \else\def\@currentlabel{\refmarkup{#1}}{\label{#3}}\fi%
    \@esphack%
    \labelmarkup{#2}
}
\makeatother

\renewcommand{\paragraph}[1]{\vspace{2pt}\noindent\textbf{#1}\hspace{1em}}

\title{Learning Dynamics Models with Stable Invariant Sets}

\author{
Naoya Takeishi\textsuperscript{\rm 1,3},
Yoshinobu Kawahara\textsuperscript{\rm 2,3}\\}

\affiliations {
    \textsuperscript{\rm 1}University of Applied Sciences and Arts Western Switzerland
    \hspace{2em}
    \textsuperscript{\rm 2}Kyushu University \\
    \textsuperscript{\rm 3}RIKEN Center for Advanced Intelligence Project \\
    naoya.takeishi@riken.jp
}

\begin{document}

\maketitle

\begin{abstract}
Invariance and stability are essential notions in dynamical systems study, and thus it is of great interest to learn a dynamics model with a stable invariant set. However, existing methods can only handle the stability of an equilibrium. In this paper, we propose a method to ensure that a dynamics model has a stable invariant set of general classes such as limit cycles and line attractors. We start with the approach by \citet{manekLearningStableDeep2019}, where they use a learnable Lyapunov function to make a model stable with regard to an equilibrium. We generalize it for general sets by introducing projection onto them. To resolve the difficulty of specifying a to-be stable invariant set analytically, we propose defining such a set as a primitive shape (e.g., sphere) in a latent space and learning the transformation between the original and latent spaces. It enables us to compute the projection easily, and at the same time, we can maintain the model's flexibility using various invertible neural networks for the transformation. We present experimental results that show the validity of the proposed method and the usefulness for long-term prediction.
\end{abstract}



\section{Introduction}
\label{intro}

Machine learning of dynamical systems appears in diverse disciplines, such as physics \citep{raissiPhysicsinformedNeuralNetworks2019}, biology \citep{costelloMachineLearningApproach2018}, chemistry \citep{liMolecularDynamicsOnthefly2015}, and engineering \citep{mortonDeepDynamicalModeling2018}.
Recent progress in dynamics models includes the Gaussian process dynamics models \citep{wangGaussianProcessDynamical2006} and models based on deep neural networks \citep[e.g.,][]{takeishiLearningKoopmanInvariant2017,luschDeepLearningUniversal2018,chenNeuralOrdinaryDifferential2018,manekLearningStableDeep2019,greydanusHamiltonianNeuralNetworks2019}.
Whatever models are employed, we often would like to know and control the nature of a learned dynamics model, for example, to reflect prior knowledge of the dynamics and to ensure specific behavior of the learned dynamics model.

\emph{Invariance} and \emph{stability} of some subsets of a state space play a key role in dynamical systems study as they concisely describe the asymptotic behavior (i.e., in $t\to\infty$) of a system.
For example, dynamical systems with stable invariant sets are used to explain neurological functions such as working memory and eye control \citep{eliasmithUnifiedApproachBuilding2005}.
Moreover, various self-sustained oscillations in physical, chemical, or biological phenomena are modeled as systems with stable closed orbits \citep{strogatzbook}.
In the dynamical systems study, \emph{analyzing} system's stability has been a classical yet challenging problem.
In contrast, \emph{synthesizing} (i.e., achieving) stability of some dynamics models is a problem that has been addressed mainly in automatic control and machine learning.
Whereas control theory has been trying to achieve stable systems by designing control inputs, another possible strategy is to learn a dynamics model from data with a constraint that the model attains some desired stability property.

In this work, we tackle the problem of learning dynamics models with guaranteed invariance and stability.
This problem is important in many practices of machine learning.
For example, we often have prior knowledge that a target phenomenon shows self-sustained oscillations \citep{strogatzbook}.
Such prior knowledge is a powerful inductive bias in learning and can be incorporated into learning by forcing a model to have a stable limit cycle.
Likewise, we often want to assure the invariance and stability (i.e., time-asymptotic behavior) of a learned forecasting model for meaningful prediction or safety issues.
To these ends, we need a method to guarantee invariance and stability of a dynamics model.

Learning dynamics models with provable stability is not a new task.
Learning linear dynamical systems with stability \citep[e.g.,][]{lacySubspaceIdentificationGuaranteed2003,siddiqiConstraintGenerationApproach2008,huangLearningStableLinear2016} is a long-standing problem, and learning stable nonlinear dynamics has also been addressed by many researchers \citep[e.g.,][]{khansari-zadehLearningStableNonlinear2011,neumannNeuralLearningStable2013,umlauftLearningStableStochastic2017,dunckerLearningInterpretableContinuoustime2019,changNeuralLyapunovControl2019,manekLearningStableDeep2019,massaroliStableNeuralFlows2020}.
However, these methods can only handle the stability of a finite number of equilibria (i.e., points where a state remains if no external perturbation applies) and are not suitable for guaranteeing general stable invariant sets (e.g., limit cycles, limit tori, and continuous sets of infinitely many equilibria).
This limitation has been hindering useful applications of machine learning on dynamical systems, for example in physics and biology.

We develop a dynamics model with provable stability of general invariant sets.
The starting point of our model is the approach by \citet{manekLearningStableDeep2019}, where they use a learnable Lyapunov function to modify a base dynamics model to ensure the stability of an equilibrium.
We generalize it for handling general subsets of state space (e.g., closed orbits, surfaces, and volumes) as stable invariant sets, by introducing projection onto such sets in the definition of the learnable Lyapunov function.
A practical difficulty arising here is that in general, we cannot specify the geometry of a to-be stable invariant set analytically.
To resolve this difficulty, we propose defining such a set as a primitive shape (e.g., a sphere) in a latent space and learning the transformation between the original state space and the latent space (see Figure~\ref{fig:diagram}).
We can configure such a primitive shape so that the projection is easily computed.
At the same time, we can maintain the flexibility of the model using the rich machinery of invertible neural networks that have been actively studied recently \citep[see, e.g.,][]{papamakarios_normalizing_2019}.

In the remainder, we review the technical background in Section~\ref{back}.
We first give a general definition of the proposed dynamics model in Section~\ref{main} and then show its concrete constructions in Section~\ref{example}.
We introduce some related studies in Section~\ref{related}.
We present experimental results in Section~\ref{expt}, with which we can confirm the validity of the proposed method and its usefulness for the application of long-term prediction.
The paper is concluded in Section~\ref{concl}.


\section{Background}
\label{back}

\subsection{Invariant Sets of Dynamical Systems}

We primarily consider a continuous-time dynamical system described by an ordinary differential equation
\begin{equation}\label{eq:ds}
    \dot{\bm{x}} = \bm{f}(\bm{x}),
\end{equation}
where $\bm{x}\in\mathcal{X}\subseteq\mathbb{R}^d$ is a state vector in a state space $\mathcal{X}$.
$\dot{\bm{x}}$ denotes the time derivative, $\mathrm{d}\bm{x}/\mathrm{d}t$.
We assume that $\bm{f}:\mathcal{X}\to\mathcal{X}$ is a locally Lipschitz function.
We denote the solution of \eqref{eq:ds} with initial condition $\bm{x}(0)=\bm{x}_0$ as $\bm{x}(t)$.
An \emph{invariant set} of a dynamical system is defined as follows:
\begin{definition}[Invariant set]
    An invariant set $\invset$ of dynamical system \eqref{eq:ds} is a subset of $\mathcal{X}$ such that a trajectory $\bm{x}(t)$ starting from $\bm{x}_0\in\invset$ remains in $\invset$, i.e., $\bm{x}(t)\in\invset$ for all $t\geq0$.
\end{definition}


\subsection{Stability of Equilibrium}

A state $\bm{x}_\mathrm{e}$ s.t. $\bm{f}(\bm{x}_\mathrm{e})=0$ is called an \emph{equilibrium} of \eqref{eq:ds} and constitutes a particular class of invariant sets.
One of the common interests in analyzing dynamical systems is the \emph{Lyapunov stability} of equilibria \citep[e.g.,][]{hsrbook,gieslReviewComputationalMethods2015}.
Informally, an equilibrium $\bm{x}_\mathrm{e}$ is stable if the trajectories starting near $\bm{x}_\mathrm{e}$ remain around it all the time.
More formally;
\begin{definition}[Stability of equilibrium]
    An equilibrium $\bm{x}_\mathrm{e}$ is said to be Lyapunov stable if for every $\epsilon>0$, there exists $\delta>0$ such that, if $\Vert\bm{x}(0)-\bm{x}_\mathrm{e}\Vert<\delta$, then $\Vert\bm{x}(t)-\bm{x}_\mathrm{e}\Vert<\epsilon$ for all $t\geq0$.
    Moreover, if $\bm{x}_\mathrm{e}$ is stable, and $\bm{x}(t)\to\bm{x}_\mathrm{e}$ as $t\to\infty$, $\bm{x}_\mathrm{e}$ is said to be \emph{asymptotically stable}.
\end{definition}
The (asymptotic) stability of equilibria plays a crucial role in analyzing dynamical systems as well as in applications.
For example, in computational neuroscience, dynamical systems with stable equilibria are used to explain phenomena such as associative memory and pattern completion.
In physics, coupled phase oscillators whose equilibria are stable are used to model synchronization phenomena.
In engineering, equilibrium stability is used for roughly assessing the safety of controlled agents and the plausibility of forecasting.

Lyapunov's direct method is a well-known way to assess the stability of equilibria \citep[see, e.g.,][]{hsrbook}, which can be summarized as follows:
\begin{theorem}[Lyapunov's direct method]\label{thm:lyap}
    Let $\bm{x}_\mathrm{e}$ be an equilibrium of dynamical system \eqref{eq:ds}.
    Let $V:\mathcal{U}\to\mathbb{R}$ be a function on a neighborhood $\mathcal{U}$ of $\bm{x}_\mathrm{e}$, and further suppose:
    \begin{enumerate}[topsep=2pt,itemsep=0pt,leftmargin=2em,label=(\Alph*)]
        \item\label{item:lyap:a} $V$ has a minimum at $\bm{x}_\mathrm{e}$; e.g., a sufficient condition is:\\ $(\forall \bm{x}\in\mathcal{U} ~~ V(\bm{x}) \geq 0) \, \land \, (V(\bm{x})=0 \Leftrightarrow \bm{x}=\bm{x}_\mathrm{e})$.
        \item\label{item:lyap:b} $V$ is strictly decreasing along trajectories of \eqref{eq:ds}; e.g., when $V$ is differentiable, a sufficient condition is:\\ $(\forall \bm{x} \in \mathcal{U} \backslash \{\bm{x}_\mathrm{e}\} ~~ \dot{V} = \mathrm{d}V/\mathrm{d}t = \langle \nabla V(\bm{x}), \bm{f}(\bm{x}) \rangle <0$.
    \end{enumerate}
    If such a function $V$ exists, then it is called a Lyapunov function, and $\bm{x}_\mathrm{e}$ is asymptotically stable.
\end{theorem}






\subsection{Dynamics Models with Stable Equilibrium}
\label{back:staeq}

\citet{manekLearningStableDeep2019} proposed a concise method to ensure the stability of an equilibrium of a dynamics model by construction.
They suggested learning a function $V$ that satisfies condition \ref{item:lyap:a} in Theorem~\ref{thm:lyap} with neural networks and projecting outputs of a base dynamics model onto a space where condition \ref{item:lyap:b} also holds.
Consequently, the modified model's equilibrium becomes asymptotically stable.

Their dynamics model, $\dot{\bm{x}}=\bm{f}(\bm{x})$, is built as
\begin{equation}\label{eq:mod_eq}\begin{gathered}
    \bm{f}(\bm{x}) =
    \begin{cases}
        \hat{\bm{f}}(\bm{x}) - \frac{\beta(\bm{x})}{\Vert \nabla V(\bm{x}) \Vert_2^2} \nabla V(\bm{x}), & \text{if} ~ \beta(\bm{x}) \geq 0, \\
        \hat{\bm{f}}(\bm{x}), & \text{otherwise},
    \end{cases} \\
    \text{where} \quad \beta(\bm{x}) = \nabla V(\bm{x})^\tr \hat{\bm{f}}(\bm{x}) + \alpha V(\bm{x}).
\end{gathered}\end{equation}
Here, $\hat{\bm{f}}:\mathbb{R}^d\to\mathbb{R}^d$ is a base dynamics model, and $\alpha\geq0$ is a nonnegative constant.
Function $V:\mathbb{R}^d\to\mathbb{R}$ works as a Lyapunov (candidate) function.
$V$ is designed so that it has a global minimum at $\bm{x}=\bm{0}$ and no local minima:
\begin{equation}\label{eq:V_eq}
    V(\bm{x}) = \sigma \left( q(\bm{x}) - q(\bm{\bm{x}}_\mathrm{e}) \right) + \varepsilon \Vert \bm{x} - \bm{x}_\mathrm{e} \Vert_2^2,
\end{equation}
where $\varepsilon>0$ is a positive constant to ensure the positivity of $V$, and $\sigma:\mathbb{R}\to\mathbb{R}_{\geq0}=[0,\infty)$ is a convex nondecreasing function with $\sigma(0)=0$.
Function $q:\mathbb{R}^d\to\mathbb{R}$ also needs to be convex, and they use the input convex neural networks \citep{amosInputConvexNeural2017} for $q$.


\section{Proposed Method}
\label{main}

\subsection{Stability of General Invariant Set}

We begin with reviewing the theory around stability of general invariant sets, which comprises the theoretical backbone of the proposed method.
First, stability of a general invariant set is formally defined as follows:
\begin{definition}[Stability of invariant set]
    Let $\invset\subseteq\mathcal{X}$ be a positively invariant set of dynamical system \eqref{eq:ds}, and let $\operatorname{dist}(\bm{x},\invset)=\inf_{\bm{s}\in\invset}\Vert\bm{x}-\bm{s}\Vert$ denote the distance between $\bm{x}$ and $\invset$.
    $\invset$ is said to be stable if for every $\epsilon>0$, there exists $\delta>0$ such that, if $\operatorname{dist}(\bm{x}(0), \invset)<\delta$, then $\operatorname{dist}(\bm{x}(t),\invset)<\epsilon$ for all $t\geq0$.
    Moreover, if $\invset$ is stable, and $\operatorname{dist}(\bm{x},\invset)\to0$ as $t\to\infty$, $\invset$ is said to be asymptotically stable.
\end{definition}
Stable invariant sets appear in various forms in a variety of dynamics.
For example, a closed invariant orbit is called a limit cycle, and if it is asymptotically stable, nearby trajectories approach to it as $t\to\infty$.
Such stable limit cycles play a key role in understanding behavior of various physical and biological phenomena with oscillation \citep{strogatzbook}.
Moreover, invariant sets comprising infinitely many equilibria are often considered in analyzing higher-order coupled oscillators in physics \citep{tanakaMultistableAttractorsNetwork2011} and continuous attractor networks in neuroscience \citep{eliasmithUnifiedApproachBuilding2005}.

The LaSalle's theorem characterizes the asymptotic stability of a general invariant set \citep[see, e.g.,][]{khalilbook}:
\begin{theorem}[LaSalle's theorem]\label{thm:lasalle}
    Let $\Omega \subseteq \mathcal{D} \subset \mathbb{R}^d$ be a compact set that is positively invariant for the dynamical system \eqref{eq:ds}.
    Let $V:\mathcal{D}\to\mathbb{R}$ be a differentiable function such that $\dot{V}(\bm{x}) \leq 0$ in $\Omega$.
    Let $\mathcal{E}\subseteq\Omega$ be the set of all points in $\Omega$ such that $\dot{V}(\bm{x})=0$.
    Let $\invset \subseteq \mathcal{E}$ be the largest invariant set in $\mathcal{E}$.
    Then, every solution of \eqref{eq:ds} starting from a point in $\Omega$ approaches to $\invset$ as $t\to\infty$.
\end{theorem}


\subsection{Dynamics Models with Stable Invariant Set}
\label{main:main}

We give a general framework to construct a dynamics model with a general stable invariant set.
This framework can be implemented with any parametric function approximators, such as neural networks, as its components.
We provide concrete examples of implementation later in Section~\ref{example}.

\begin{figure}[t]
    \centering
    \includegraphics[clip,width=0.98\linewidth]{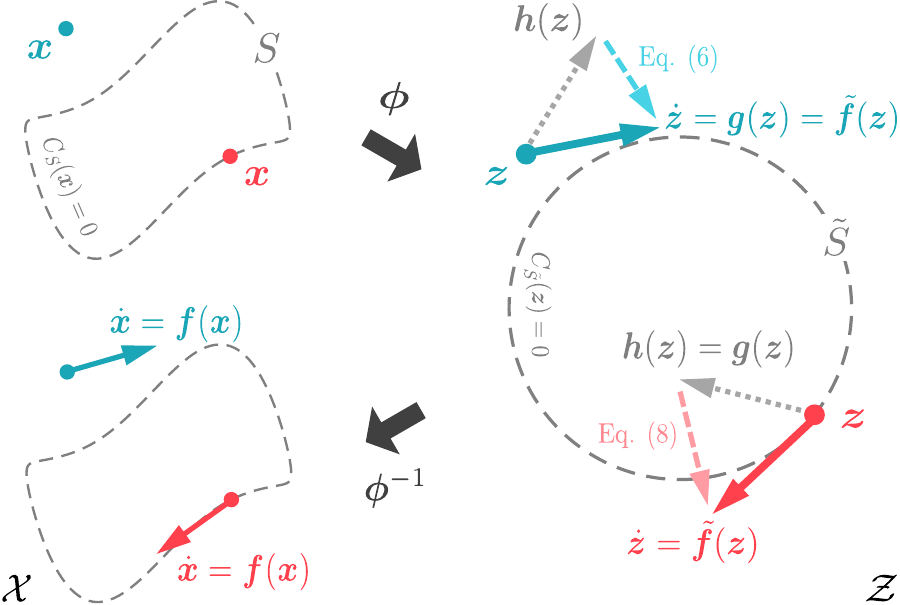}
    \caption{Proposed dynamics model, $\dot{\bm{x}}=\bm{f}(\bm{x})$, for the case where latent stable invariant set $\tilde{\invset}$ is defined as $\tilde{\invset}^\text{surf}$ in \eqref{eq:invset:surf}. Two states, $\bm{x}\notin\invset$ (blue) and $\bm{x}\in\invset$ (red), are shown. The dotted orbits are the original and latent stable invariant sets, $\invset\in\mathcal{X}$ and $\tilde{\invset}\in\mathcal{Z}$. Input $\bm{x}\in\mathcal{X}$ is first transformed into a latent state $\bm{z}\in\mathcal{Z}$ by a learnable bijective function $\bm\phi$ and then fed into a base dynamics model $\bm{h}$. $\bm{h}(\bm{z})$ is modified to ensure the stability and/or invariance of $\tilde{\invset}$ by \eqref{eq:mod1} and/or \eqref{eq:mod2}, respectively. Finally, things are projected back to $\mathcal{X}$ by $\bm\phi^{-1}$.}
    \label{fig:diagram}
\end{figure}

The proposed dynamics model, $\dot{\bm{x}}=\bm{f}(\bm{x})$, is depicted in Figure~\ref{fig:diagram}.
It comprises five steps.
Given a state vector $\bm{x}$ as an input, it first computes a transformed latent state $\bm{z}=\bm\phi(\bm{x})$ via a learnable bijective function $\bm\phi$ (\textbf{Step~1}).
Latent state $\bm{z}$ is fed into a base dynamics model $\bm{h}$ (\textbf{Step~2}).
Then, $\bm{h}(\bm{z})$ may be modified to be $\bm{g}(\bm{z})$ to ensure stability of some set $\tilde\invset$ (\textbf{Step~3}).
$\bm{g}(\bm{z})$ may further be modified to be $\tilde{\bm{f}}(\bm{z})$ to ensure invariance of $\tilde\invset$ (\textbf{Step~4}).
Finally, it computes $\bm{f}(\bm{x})=\bm\phi^{-1}(\tilde{\bm{f}}(\bm{x}))$ via the inverse of $\bm\phi$ (\textbf{Step~5}).
In the following, we explain the details of the five steps.


\subsubsection{Step~1: Learnable Invertible Feature Transform}

Given a state vector $\bm{x}\in\mathcal{X}\subseteq\mathbb{R}^d$ as an input, we transform it into a latent state $\bm{z}\in\mathcal{Z}\subseteq\mathbb{R}^{d}$ using a learnable \emph{bijective} function $\bm\phi:\mathcal{X}\to\mathcal{Z}$, that is,
\begin{equation}\label{eq:tran}
    \bm{z} = \bm\phi(\bm{x}).
\end{equation}
We restrict $\bm\phi$ to be bijective for \emph{provable} existence of a stable invariant set.
A bijective function $\bm\phi$ can be modeled in a number of ways given the development of invertible neural networks in the area of normalizing flows \citep[see, e.g.,][and references therein]{papamakarios_normalizing_2019}.
Hence, we believe that restricting $\bm\phi$ to be bijective does not severely limit the flexibility of the proposed dynamics model.

Such a feature transform, together with its inverse in \textbf{Step~5}, is indispensable when we cannot exactly parametrize the geometry of a \emph{to-be} stable invariant set (in $\mathcal{X}$) that a learned dynamics model should have.
This is often the case in practice; for example, we may only know the presence of a limit cycle but cannot describe its shape analytically in advance.
Our proposal lies in avoiding such a difficulty by defining a set that has a primitive shape (e.g., a unit sphere) in $\mathcal{Z}$, expecting (inverse of) $\bm\phi$ should learn an appropriate transformation of the primitive shape in $\mathcal{Z}$ into a to-be stable invariant set in the original space, $\mathcal{X}$.
Hereafter, a to-be stable invariant set in $\mathcal{X}$ and the corresponding primitive set in $\mathcal{Z}$ are denoted by $\invset\subseteq\mathcal{X}$ and $\tilde{\invset}\subseteq\mathcal{Z}$, respectively.


\subsubsection{Step~2: Base Dynamics Model}

The second component is a base dynamics model $\bm{h}:\mathcal{Z}\to\mathcal{Z}$ that acts on the latent state $\bm{z}$.
We can use any parametric models as $\bm{h}$.
Note that we do not have control on the invariance and stability properties of base dynamics model $\dot{\bm{z}}=\bm{h}(\bm{z})$.
Hence, we need to modify the output of $\bm{h}$ to make $\tilde{\invset}$ a stable invariant set.

\subsubsection{Step~3: Ensuring Stability}

In this step, we modify the output of the base dynamic model, $\bm{h}$, so that $\bm{z}$'s trajectories converge to some limit set $\tilde{\invset}\subseteq\mathcal{Z}$ in $t\to\infty$.
According to Theorem~\ref{thm:lasalle}, for trajectories to converge to $\tilde{\invset}$, it is sufficient that there exists a function $V:\mathcal{Z}\to\mathbb{R}$ whose value decreases along trajectories everywhere outside $\tilde{\invset}$.
To this end, we construct a candidate function for $V$ by generalizing the method of \citet{manekLearningStableDeep2019}.

Suppose $\tilde{\invset}\subseteq\mathcal{Z}$ is convex.
This assumption does not make us lose much generality because even if $\tilde{\invset}$ is convex, corresponding $S\subseteq\mathcal{X}$ is not necessarily convex thanks to the feature transform $\bm\phi$ in \textbf{Step~1} and \textbf{Step~5}.
Let $\mathbb{P}_{\tilde{\invset}}\bm{z}$ denote the orthogonal projection of $\bm{z}$ onto $\tilde{\invset}$, that is, $\mathbb{P}_{\tilde{\invset}}\bm{z} = \arg{\textstyle \min_{\bm{s}\in\tilde{\invset}}} ~ \Vert \bm{z} - \bm{s} \Vert_2^2$.
Let $q:\mathcal{Z}\to\mathbb{R}$ be a convex function, and $\sigma:\mathbb{R}\to\mathbb{R}_{\geq0}$ be a convex nonnegative nondecreasing function with $\sigma(0)=0$.
We define a function $V$ as
\begin{equation}\label{eq:V}
    V(\bm{z}) = \sigma \left( q(\bm{z}) - q\big( \mathbb{P}_{\tilde{\invset}}\bm{z} \big) \right) + \varepsilon \Vert \bm{z} - \mathbb{P}_{\tilde{\invset}}\bm{z} \Vert_2^2,
\end{equation}
with $\varepsilon>0$.
It reaches the minimum $V(\bm{z})=0$ at $\bm{z}\in\tilde{\invset}$ and does not have any local minima at $\bm{z}\notin\tilde{\invset}$ from construction.
Given such a function \eqref{eq:V}, we modify the outputs of the base dynamics model, $\bm{h}(\bm{z})$, into $\bm{g}(\bm{z})$ as follows:
\begin{equation}\label{eq:mod1}\begin{gathered}
    \bm{g}(\bm{z}) \!=\!
    \begin{cases}
        \bm{h}(\bm{z}), & \bm{z} \in \tilde{\invset}, \\
        \bm{h}(\bm{z}) - \stepfunc \big( \beta(\bm{z}) \big) \frac{\beta(\bm{z}) + \eta(\bm{z})}{\nablanorm{V}{\bm{z}}} \nabla V(\bm{z}), & \bm{z} \notin \tilde{\invset},
    \end{cases}\\
    \text{where}\quad\beta(\bm{z}) = \lyapcond{\bm{h}}{\bm{z}}.
\end{gathered}\end{equation}
Here, $\stepfunc$ is the unit step function (i.e., $\stepfunc(a)=1$ if $a\geq0$ and $\stepfunc(a)=0$ otherwise), and $\alpha\geq0$ is a nonnegative constant.
$\eta:\mathbb{R}^d\to\mathbb{R}_{\geq0}$ is a nonnegative function that works like a slack variable.
Merely setting $\eta(\bm{z})=0$ also ensures the stability of $\tilde{\invset}$, but it may be useful to make $\eta$ a learnable component if we want more flexibility.
Note that \eqref{eq:V} and \eqref{eq:mod1} do not ensure anything about the positive invariance property of $\tilde{\invset}$; it is deferred in \textbf{Step~4}.

Comparing \eqref{eq:V} and \eqref{eq:mod1} to \eqref{eq:V_eq} and \eqref{eq:mod_eq}, we can see that \textbf{Step~3} here is indeed a generalized version of the method of \citet{manekLearningStableDeep2019}.
We note that such a generalization is meaningful only with other components of the proposed method; we need the learnable feature transform of \textbf{Step~1} and \textbf{Step~5} to avoid difficulty of parametrizing a stable invariant set analytically, and the procedure of \textbf{Step~4} is indispensable to ensure that trajectories do not escape from a limit set.
\textbf{Step~3} does not work without these remedies.

We may compute $\mathbb{P}_{\tilde{\invset}}$ in a closed form when $\tilde{\invset}$ is a simple-shaped set like a sphere or a 2-torus.
Such a simple $\tilde{\invset}$ does not severely drop the flexibility of a dynamics model if we set $\bm\phi$ to be flexible enough.
Meanwhile, we can also adopt $\tilde{\invset}$ with nontrivial $\mathbb{P}_{\tilde{\invset}}$ if needed, by employing the technique of the convex optimization layer \citep{agrawalDifferentiableConvexOptimization2019}.


\subsubsection{Step~4: Ensuring Invariance}

Recall that the previous step only ensures $\tilde{\invset}$ is a limit set.
Even if trajectories converge to $\tilde{\invset}$ in $t\to\infty$, they may escape unless it is also invariant.
To make $\tilde{\invset}$ invariant, we further modify the output of $\bm{g}$.

Without loss of generality, we consider the following two types of the definition of $\tilde{\invset}$:
\begin{subequations}\label{eq:invset}\begin{align}
    \label{eq:invset:vol}
    \tilde{\invset}^\text{vol} &= \{ \bm{z} \mid C_{\tilde{\invset}}(\bm{z}) \geq 0 \}, \\
    \label{eq:invset:surf}
    \tilde{\invset}^\text{surf} &= \{ \bm{z} \mid C_{\tilde{\invset}}(\bm{z}) = 0 \},
\end{align}\end{subequations}
where $C_{\tilde{\invset}}:\mathcal{Z}\to\mathbb{R}$ is a continuously differentiable function.
The invariance of such sets can be characterized as follows (a proof is in the appendix):
\begin{proposition}\label{prop:inv}
    For a dynamical system $\dot{\bm{z}}=\bm{F}(\bm{z})$ with some $\bm{F}:\mathcal{Z}\to\mathcal{Z}$,
    \begin{enumerate}[topsep=2pt,itemsep=0pt,leftmargin=2em,label=(\alph*)]
        \item\label{item:prop:inv:a} If $C_{\tilde{\invset}}(\bm{z}) = 0 \, \Rightarrow \, \surfcond{\bm{F}}{\bm{z}} > 0$, then $\tilde{\invset}^\text{vol}$ in \eqref{eq:invset:vol} is a positively invariant set.
        \item\label{item:prop:inv:b} If $C_{\tilde{\invset}}(\bm{z}) = 0 \, \Rightarrow \, \surfcond{\bm{F}}{\bm{z}} = 0$, then $\tilde{\invset}^\text{surf}$ in \eqref{eq:invset:surf} is a positively invariant set.
    \end{enumerate}
\end{proposition}
Given this fact, we modify the outputs of the previous step, $\bm{g}(\bm{z})$, into $\tilde{\bm{f}}(\bm{z})$ as follows:
\begin{equation}\label{eq:mod2}\begin{gathered}
    \tilde{\bm{f}}(\bm{z}) =
    \begin{cases}
        \bm{g}(\bm{z})
        - \frac{\gamma(\bm{z}) - \xi(\bm{z})}{\nablanorm{C_{\tilde{\invset}}}{\bm{z}}} \nabla C_{\tilde{\invset}}(\bm{z}), & C_{\tilde{\invset}}(\bm{z}) = 0, \\
        \bm{g}(\bm{z}), & C_{\tilde{\invset}}(\bm{z}) \neq 0,
    \end{cases} \\
    \text{where}\quad\gamma(\bm{z}) = \surfcond{\bm{g}}{\bm{z}}.
\end{gathered}\end{equation}
The definition of $\xi$ depends on that of $\tilde{\invset}$; if $\tilde{\invset}$ is defined as $\tilde{\invset}^\text{vol}$ in \eqref{eq:invset:vol}, $\xi:\mathcal{Z}\to\mathbb{R}_{>0}$ is a positive-valued function; if $\tilde{\invset}$ is as $\tilde{\invset}^\text{surf}$ in \eqref{eq:invset:surf}, it is simply $\xi(\bm{z})=0$.
Note that in actual computation, condition $C_{\tilde{\invset}}(\bm{z})=0$ in \eqref{eq:mod2} should be replaced by $\vert C_{\tilde{\invset}}(\bm{z}) \vert \leq \epsilon$ with a tiny $\epsilon$.


\subsubsection{Step~5: Projecting Back}

Things have been described in terms of the latent state $\bm{z}\in\mathcal{Z}$ after \textbf{Step~1}.
However, what we want is a dynamics model on $\bm{x}\in\mathcal{X}$, namely $\dot{\bm{x}}=\bm{f}(\bm{x})$.
As the final part of the proposed method, we project things back to $\mathcal{X}$ via the inverse of $\bm\phi$, that is,
\begin{equation}\label{eq:final}
    \bm{f}(\bm{x}) = \bm\phi^{-1} \big( \bm{\tilde{f}}(\bm{z}) \big) = \bm\phi^{-1} \Big( \bm{\tilde{f}} \big( \bm\phi(\bm{x}) \big) \Big).
\end{equation}
Recall that we assumed $\bm\phi$ is invertible in \textbf{Step~1}.


\subsection{Analysis}
\label{main:analysis}

The dynamics model in \eqref{eq:final} has a stable invariant set that can be learned from data.
We summarize such a property as follows, where we describe the cases of $\cdot^\text{vol}$ and $\cdot^\text{surf}$ in parallel.
A proof is found in the appendix.
\begin{proposition}\label{prop:main}
    Let $\tilde\invset^\text{vol}$ (or $\tilde\invset^\text{surf}$) be a subset of $\mathcal{Z}\subseteq\mathbb{R}^{d}$ defined in \eqref{eq:invset:vol} (or \eqref{eq:invset:surf}).
    Let $\tilde{\bm{f}}:\mathcal{Z}\to\mathcal{Z}$ be the function in \eqref{eq:mod2}.
    Then, for a dynamical system $\dot{\bm{z}}=\tilde{\bm{f}}(\bm{x})$, $\tilde\invset^\text{vol}$ (or $\tilde\invset^\text{surf}$) is a positively invariant set and is asymptotically stable.
\end{proposition}
\begin{corollary}\label{coro:main}
     Suppose a subset of $\mathcal{X}$, namely $\invset^\text{vol} = \{ \bm{x} \mid C_{\tilde{\invset}}(\bm\phi(\bm{x})) \geq 0 \}$ (or $\invset^\text{surf} = \{ \bm{x} \mid C_{\tilde{\invset}}(\bm\phi(\bm{x})) = 0 \}$), where $C_{\tilde{\invset}}$ is the function that appeared in \eqref{eq:invset}.
     Let $\bm{f} = \bm\phi^{-1} \circ \tilde{\bm{f}} \circ \bm\phi$ as in \eqref{eq:final}.
     Then, $\invset^\text{vol}$ (or $\invset^\text{surf}$) is an asymptotically stable invariant set of a dynamical system defined as $\dot{\bm{x}}=\bm{f}(\bm{x})$.
\end{corollary}


\subsection{Extension}

The proposed dynamics model in Section~\ref{main:main} works not only as a standalone machine learning model, but also as a module embedded in a larger machine learning method.
For example, suppose we have high-dimensional observations $\bm{y}\in\mathcal{Y}$ (e.g., observation of fluid flow).
In such a case, we often try to transform $\bm{y}$ into lower-dimensional vectors, namely $\bm{x}$, using methods like principal component analysis and autoencoders.
We can then consider a dynamics model (with a stable invariant set) on $\bm{x}$ as in Section~\ref{main:main}, rather than on $\bm{y}$ directly.
Temporal forecasting is performed in the space of $\bm{x}$ and then returned to the space of $\bm{y}$.

Such an extension is straightforward yet useful, but a drawback is that if the dimensionality reduction is lossy, which is often the case, we can no longer guarantee a stable invariant set in $\mathcal{Y}$.
Nonetheless, such an approximative model may still be useful.
We exemplify such a case in Section~\ref{expt:fluid}, where we reduce the dimensionality of fluid flow observations by principal component analysis and learn a dynamics model on the low-dimensional space.


\section{Implementation Examples}
\label{example}

\subsection{Learnable Components}

The components of the proposed method, namely $\bm\phi$, $\bm{h}$, $q$, $\eta$, and $\xi$, can be any parametric models such as neural networks.
Let us introduce examples for each component.

\paragraph{$\bm\phi$}
Choice of $\bm\phi$'s model depends on the availability of prior knowledge of the dynamics to be learned (see also Table~\ref{tab:example}).
For example, if we know the topology of $\invset$ (e.g., it is a closed orbit), we can model $\bm\phi$ as a diffeomorphic function such as the neural ODE (NODE) \citep{chenNeuralOrdinaryDifferential2018}.
In fact, such prior knowledge is often available from our scientific understanding of physical, chemical, and biological phenomena \citep[see, e.g.,][]{strogatzbook,tanakaMultistableAttractorsNetwork2011,eliasmithUnifiedApproachBuilding2005}.
We may use other types of invertible models if less prior knowledge is available; for example, a neural ODE with auxiliary variables (namely, ANODE) \citep{dupontAugmentedNeuralODEs2019} can represent non-homeomorphic functions. 
If perfect prior knowledge is available (i.e., we can specify the geometry of $\invset\subseteq\mathcal{X}$ analytically), simply set $\bm{z}=\bm\phi(\bm{x})=\bm{x}$.

\paragraph{$\bm{h}$}
We can substitute arbitrary models to the base dynamics, $\bm{h}$, in accordance with the nature of data.

\paragraph{$q$}
The convex function, $q$ in \eqref{eq:V}, can be modeled using the input-convex neural networks \citep{amosInputConvexNeural2017} as in the previous work \citep{manekLearningStableDeep2019}.

\paragraph{$\eta$ and $\xi$}
The slack-like functions, $\eta$ and $\xi$ in \eqref{eq:mod1} and \eqref{eq:mod2}, respectively, can be modeled as neural networks with output-value clipped to be nonnegative or positive.


\subsection{Stable Invariant Set}
\label{example:invset}

Besides the learnable components, we should prepare a to-be stable invariant set.
If we do not know the analytic form of $\invset\in\mathcal{X}$ (i.e., $C_\invset$), which is usually the case, we are to define $\tilde{\invset}\in\mathcal{Z}$ (i.e., $C_{\tilde{\invset}}$) instead.
A general guideline we suggest is to set $\tilde{\invset}$ as a simple primitive shape, such as spheres and tori.
As stated earlier, setting $\tilde{\invset}$ to be a primitive shape does not severely restrict the flexibility of the model, thanks to the learnable feature transform, $\bm\phi$, which ``deforms'' a simple $\tilde{\invset}$ to various $\invset$.
We can regard unknown coefficients in $C_{\tilde{\invset}}$ (e.g., radius of sphere) as learnable parameters, too.

We can also consider the case of low-dimensional $\invset$ by setting $\tilde{\invset}$ also low-dimensional.
Here, axes of $\mathcal{Z}$ ignored by a low-dimensional $\tilde{\invset}$ can be arbitrary because the feature transform $\bm\phi$ modeled by a neural network is usually flexible enough to learn a rotation between $\mathcal{Z}$ and $\mathcal{X}$.

Care may have to be taken in the computation of $\mathbb{P}_{\tilde{\invset}}$.
If we do not know a closed form of $\mathbb{P}_{\tilde{\invset}}\bm{z}$, as long as $\tilde{\invset}$ is convex as we assumed, we can use the differentiable convex optimization layer \citep{agrawalDifferentiableConvexOptimization2019} to allow gradient-based optimization.
For example, suppose we have $\tilde{\invset}$ of the type of \eqref{eq:invset:vol}.
Then, $\mathbb{P}_{\tilde{\invset}}$ is an optimization problem:
\begin{equation}\label{eq:opt}
    \mathbb{P}_{\tilde{\invset}}\bm{z} = \arg\min_{\bm{s}} ~ \Vert \bm{z} - \bm{s} \Vert_2^2
    \quad\text{s.t.}\quad
    C_{\tilde{\invset}}(\bm{s}) \geq 0.
\end{equation}
The derivative of its output (i.e., $\partial \mathbb{P}_{\tilde{\invset}}\bm{z} / \partial \bm{z}$) can be computed by the techniques of \citet{agrawalDifferentiableConvexOptimization2019} via the implicit function theorem on the optimality condition of \eqref{eq:opt}.

\paragraph{Examples}
Let us provide concrete examples of the configuration of $\bm\phi$ and $C_{\tilde{\invset}}$ (also summarized in Table~\ref{tab:example}).
\begin{example}\label{exmpl:sphere1}
    If we exactly know that $\invset$ is a sphere around the origin, we can set $\bm\phi$ to be the identity function, i.e., $\bm{z}=\bm\phi(\bm{x})=\bm{x}$, and set $C_{\tilde{\invset}}(\bm{z}) = \Vert \bm{z} \Vert^2 - r^2$.
    In this case, $\mathbb{P}_{\tilde{\invset}}\bm{z} = \mathbb{P}_{\invset}\bm{x} = r \bm{x} / \Vert \bm{x} \Vert$ for $\bm{x} \neq \bm{0}$ (and arbitrary for $\bm{x}=\bm{0}$).
    Radius $r$ may or may not be a learnable parameter.
\end{example}
%
%
\begin{example}\label{exmpl:lc}
    If we know the dynamics should have a stable limit cycle, we can set $\invset$ to be a circle along a pair of axes of $\mathcal{Z}$, expecting $\bm\phi$ learns an appropriate coordinate transform to adjust the axes to those in the original space. For example, $C_{\tilde{\invset}}(\bm{z})=z_1^2+z_2^2-r^2$ (and ignore $z_i$'s for $i>2$).
\end{example}
\begin{example}\label{exmpl:torus}
    We may set $\tilde{\invset}$ to be a 2-torus, $C_{\tilde{\invset}}(\bm{z})=(\sqrt{z_1^2+z_2^2}-R)^2 + z_3^2 - r^2$, onto which the orthogonal projection $\mathbb{P}_{\tilde{\invset}}\bm{z}$ can be computed analytically.
\end{example}
\begin{example}\label{exmpl:plane}
    Another common option is a hyperplane $C_{\tilde{\invset}}(\bm{z})=\bm{c}^\tr\bm{z}-b$.
    This is useful in modeling, for example, sets of infinitely many equilibria, which often appear in computational neuroscience \citep{eliasmithUnifiedApproachBuilding2005}.
\end{example}
\begin{example}\label{exmpl:quadric}
    More generally, we may set $\tilde{\invset}$ to be a quadric, $C_{\tilde{\invset}}(\bm{z})=\bm{z}^\tr\bm{Q}\bm{z}+\bm{p}^\tr\bm{z}+r$.
    In this case, we need the differentiable optimization layer \citep{agrawalDifferentiableConvexOptimization2019}.
\end{example}

\begin{table}[t]
    \centering\vspace*{1.5mm}
    \setlength{\tabcolsep}{4pt}
    {\small\begin{tabular}{m{4.7cm}cc}
        \toprule
         & $\bm\phi(\bm{x})$ & $C_{\tilde{\invset}}(\bm{z})$ \\
         \midrule
         \makecell[l]{Perfect knowledge available\\[-3pt]{\scriptsize (i.e., exact $\invset\in\mathcal{X}$ is known)}} & identity & \makecell{$\tilde{\invset}=\invset$\\[-3pt]{\scriptsize (cf. Example~\ref{exmpl:sphere1})}} \\
         \midrule
         \makecell[l]{Partial knowledge available\\[-3pt]{\scriptsize (i.e., rough behavior of phenomenon is known)}} & & \\
         \makecell[r]{e.g., self-sustained oscillations} & NODE & Example~\ref{exmpl:lc} \\
         \makecell[r]{e.g., quasiperiodic patterns} & NODE & Example~\ref{exmpl:torus} \\
         \makecell[r]{e.g., neural integrators} & (A)NODE & Example~\ref{exmpl:plane} \\
         \bottomrule
    \end{tabular}}
    \caption{Implementation examples in accordance with availability of prior knowledge. NODE \citep{chenNeuralOrdinaryDifferential2018} and ANODE \citep{dupontAugmentedNeuralODEs2019} are mentioned here, but other invertible neural nets are applicable, too. For concrete examples of each case of ``partial knowledge,'' e.g., \citet{strogatzbook} and \citet{eliasmithUnifiedApproachBuilding2005} are informative.}
    \label{tab:example}
\end{table}


\subsection{Learning Procedures}
\label{example:learn}

Given a dataset and a dynamics model $\dot{\bm{x}}=\bm{f}(\bm{x})$ constructed as above, we are to learn the parameters of the unknown functions, $\bm\phi$, $\bm{h}$, and $q$, and possibly $\bm\eta$, $\bm\xi$, and $C$.
The learning scheme can be designed in either or both of the following two ways.
First, if we have paired observations of $\bm{x}$ and $\dot{\bm{x}}$, we simply minimize some loss (e.g., square loss) between $\dot{\bm{x}}$ and $\bm{f}(\bm{x})$.
This is also applicable when we can estimate $\dot{\bm{x}}$ from $\bm{x}$'s \citep[e.g.,][]{chartrandNumericalDifferentiationNoisy2011}.
Second, if we have unevenly-sampled sequences $(\bm{x}_{t_1}, \dots, \bm{x}_{t_n})$, we utilize the adjoint state method or backpropagation in forward ODE solvers for optimization \citep[see, e.g.,][]{chenNeuralOrdinaryDifferential2018}.


\section{Related Work}
\label{related}

\paragraph{Learning Stable Dynamics}
Learning stable \emph{linear} dynamical systems, e.g., $\bm{x}_{t+1}=\bm{A}\bm{x}_t$ s.t. $\rho(\bm{A})<1$, is indeed a nontrivial problem and has been addressed for decades \citep[e.g.,][]{lacySubspaceIdentificationGuaranteed2003,siddiqiConstraintGenerationApproach2008,huangLearningStableLinear2016}.
The problem of learning stable nonlinear systems has also been studied for various models, such as Gaussian mixtures \citep{khansari-zadehLearningStableNonlinear2011,blocherLearningStableDynamical2017,umlauftLearningStableStochastic2017}, Gaussian processes \citep{dunckerLearningInterpretableContinuoustime2019}, and neural networks \citep{neumannNeuralLearningStable2013,manekLearningStableDeep2019,tuorConstrainedNeuralOrdinary2020,massaroliStableNeuralFlows2020}.
However, they only handle the stability of finite number of equilibria (or usually only
an equilibrium).

\paragraph{Learning Stabilizing Controllers}
Another related direction is to learn a controller that stabilizes a given dynamical system.
For example, \citet{changNeuralLyapunovControl2019} proposed a method to learn neural controllers by constructing a neural Lyapunov function simultaneously.
They adopt a self-supervised learning scheme where a neural controller and a neural Lyapunov function are trained so that the violation of the stability condition (in Theorem~\ref{thm:lyap}) is minimized.
Such an approach is also applicable to dynamics learning, but existing methods only focus on discrete equilibria, too.

\paragraph{Learning Physically Meaningful Systems}
Another related thread of studies is to learn physical models, such as Lagrangian \citep{lutterDeepLagrangianNetworks2019,cranmerLagrangianNeuralNetworks2020} and Hamiltonian \citep{greydanusHamiltonianNeuralNetworks2019} neural networks.
They are conservative systems so do not involve stability, but an extension to port-Hamiltonian systems \citep{zhongDissipativeSymODENEncoding2020} can consider dissipative systems.
However, such a method cannot control the properties of limit sets.



\section{Experiment}
\label{expt}

\subsection{Configuration}

\paragraph{Implementation}
We implemented the learnable components (i.e., $\bm\phi$, $\bm{h}$, $q$, $\eta$, and $\xi$) with neural networks.
We used ANODE \citep{dupontAugmentedNeuralODEs2019} for $\bm\phi$ in Sections~\ref{expt:vdp} and \ref{expt:fluid} to allow much flexibility, while non-augmented NODE \citep{chenNeuralOrdinaryDifferential2018} was also sufficient.
For the other components, we used networks with fully-connected hidden layers.
We used the exponential linear unit as the activation function.
Other details are found in the appendix.

\paragraph{Baselines}
Besides the proposed model in \eqref{eq:final}, we tried either or both of the following models as baselines:
\begin{enumerate}[topsep=2pt,itemsep=0pt,leftmargin=2em,label=\arabic*)]
    \item Base dynamics model \emph{without stability nor invariance}, i.e., $\dot{\bm{x}} = \bm\phi^{-1}( \bm{h}( \bm\phi(\bm{x}) ) )$; we may refer to this baseline as a \textbf{vanilla model}.
    \item Stable dynamics model like ours, but \emph{the stable invariant set is fixed to be an equilibrium} at $\bm{x}=0$ (i.e., almost the same with \citet{manekLearningStableDeep2019}); we may refer to this baseline as a \textbf{stable equilibrium model}.
\end{enumerate}


\subsection{Simple Examples}
\label{expt:simple}

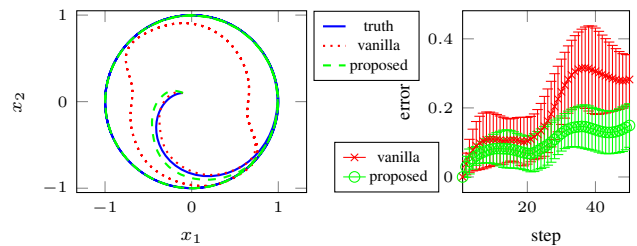
\begin{figure}
    \centering
    \begin{minipage}[t]{0.51\linewidth}
        \centering\vspace*{0pt}
        \pgfplotsset{height=4cm}
        \begin{tikzpicture}
            \begin{axis}[compat=newest,
                label style={font=\scriptsize}, xlabel={$x_1$}, ylabel={$x_2$},
                axis equal,
                enlarge x limits=false,
                xmin=-1, xmax=1, ymin=-1.05, ymax=1.05,
                xtick distance=1, ytick distance=1,
                legend pos=outer north east,
                legend entries={truth,vanilla,proposed},
                legend style={nodes={scale=0.6, transform shape}},
                ]
                \addplot [color=blue,thick] table {circsurf/truth-circsurf_gi0_traj.txt};
                \addplot [color=red,dotted,thick] table {circsurf/d1=64-64_nosta_noinv_gi0_traj.txt};
                \addplot [color=green,dashed,thick] table {circsurf/d1=64-64_d3=16_d4=0_gi0_traj.txt};
            \end{axis}
        \end{tikzpicture}
    \end{minipage}
    \begin{minipage}[t]{0.47\linewidth}
        \centering\vspace*{1.2mm}
        \pgfplotsset{height=4cm,width=3.8cm}
        \begin{tikzpicture}
            \begin{axis}[compat=newest,
                label style={font=\scriptsize}, xlabel={step}, ylabel={error},
                enlarge x limits=false,
                legend entries={vanilla,proposed},
                legend style={at={(-0.14,0)}, anchor=south east, nodes={scale=0.6, transform shape}},
                ]
                \addplot+ [red, mark=x, error bars/.cd, y dir=both, y explicit] table [x index=0, y index=1, y error index=2] {circsurf/d1=64-64_nosta_noinv_test_difabs_stat.txt};
                \addplot+ [green, mark=o, error bars/.cd, y dir=both, y explicit] table [x index=0, y index=1, y error index=2] {circsurf/d1=64-64_d3=16_d4=0_test_difabs_stat.txt};
            \end{axis}
        \end{tikzpicture}
    \end{minipage}
    \caption{Test results on the system with limit cycle in Section~\ref{expt:simple}. (\emph{left}) Examples of long-term prediction from $x_1,x_2=-.1,.1$ for $200$ steps. (\emph{right}) Average (and stdev) long-term prediction errors against prediction steps.}
    \label{fig:circsurf}
\end{figure}
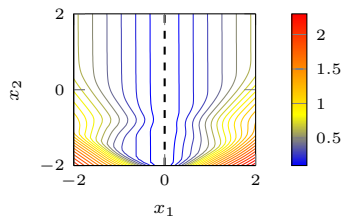
\begin{figure}
    \centering\vspace*{0pt}
    \begin{minipage}[t]{0.55\linewidth}
        \centering\vspace*{0pt}
        \pgfplotsset{height=3.6cm,width=4cm}
        \begin{tikzpicture}
            \begin{axis}[compat=newest,view={0}{90},
                    colorbar, colorbar style={width=2mm,},
                    label style={font=\scriptsize}, xlabel={$x_1$}, ylabel={$x_2$},
                ]
                \addplot3 [mesh/rows=50,mesh/num points=2500, contour gnuplot={number=25,output point meta=rawz,labels=false}] table {linear/d1=64-64_d3=16_d4=0_lyap_Vvalue.txt};
                \addplot [black,dashed,thick] coordinates {(0,-2) (0,2)};
            \end{axis}
        \end{tikzpicture}
    \end{minipage}
    \hspace{0.02\linewidth}
    \begin{minipage}[t]{0.41\linewidth}
        \centering\vspace*{1.5mm}
        \caption{Contour plot of $V(\bm{x})$ learned on the data generated from the system with line attractor in Section~\ref{expt:simple}. The dotted line is the true line attractor, $x_1=0$.}
        \label{fig:linear}
    \end{minipage}
\end{figure}

As a proof of concept, we examined the performance of the proposed method on simple dynamical systems whose stable invariant set is known analytically.
Hence, we do not need $\bm\phi$ (i.e., set $\bm\phi(\bm{x})=\bm{x}$) in the two experiments in this section.

\paragraph{Limit Cycle}
We examined the system with a limit cycle:
\begin{equation*}
    \dot{x}_1 = x_1 - x_2 - x_1(x_1^2 + x_2^2),\quad
    \dot{x}_2 = x_1 + x_2 - x_2(x_1^2 + x_2^2),
\end{equation*}
whose orbits approach to a unit circle as $t\to\infty$.
We generated four sequences of length 20 with $\Delta t = .075$ and used the pairs of $\bm{x}$ and $\dot{\bm{x}}$ as training data.
For the proposed model, we set $C_{\invset}(\bm{x})=x_1^2+x_2^2-1$ (i.e., the truth) with $S$ defined as in \eqref{eq:invset:surf}.

In Figure~\ref{fig:circsurf}, we show the results of long-term prediction given only $\bm{x}_{t=0}$ that was \emph{not in the training data}.
The left panel depicts trajectories of length $200$ predicted by the true system, the vanilla model, and the proposed model.
The proposed model's trajectory successfully reaches the plausible limit cycle, while it is not surprising as a natural consequence of the model's construction.
The right panel shows the average long-term prediction errors against prediction steps (a single step corresponds to the $\Delta t$).
The average was taken with regard to 20 test sequences with different $\bm{x}_{t=0}$.
We can observe that the proposed stable model achieves consistently lower prediction errors.

\paragraph{Line Attractor}
We examined another simple system:
\begin{equation*}
    \dot{x}_1 = x_1(1 - x_2),\quad
    \dot{x}_2 = x_1^2.
\end{equation*}
Line $x_1=0$ constitutes a line attractor of this system as a set of infinitely many stable equilibria; every orbit starting at $x_1\neq0$ approaches to some point on this line as $t\to\infty$.
We generated eight sequences of length $80$ with $\Delta t=.05$ as training data.
We learned the proposed model with $\bm\phi(\bm{x})=\bm{x}$ and $C_{\invset}(\bm{x})=c_1 x_1 + c_2 x_2$, where $c_1$ and $c_2$ were learnable parameters, and $S$ was defined as in \eqref{eq:invset:surf}.

In Figure~\ref{fig:linear}, we show the values of learned $V(\bm{x})$.
We can say it is successfully learned because $V(\bm{x})$ monotonically decreases toward the line $x_1=0$.
Moreover, it reflects the fact that a state of this system moves faster when $\vert x_1 \vert \gg 0$ and $x_2 \ll 0$ (i.e., in the lower part of the $\bm{x}$-plane).


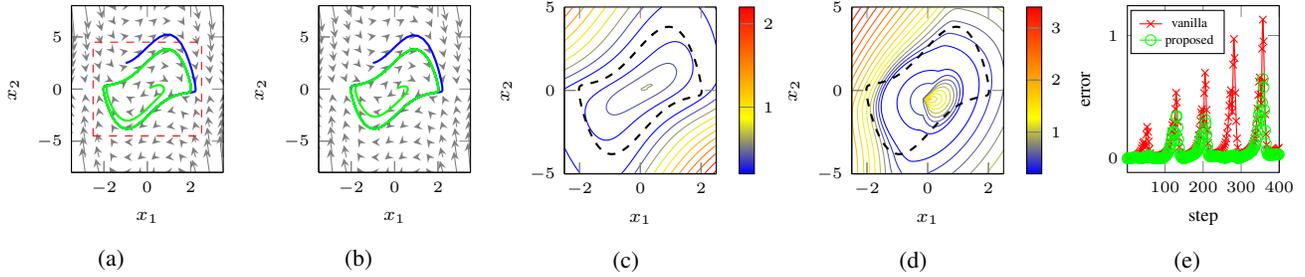
\begin{figure*}[t]
    \centering
    \begin{minipage}[t]{0.17\linewidth}
        \centering\vspace*{0pt}
        \pgfplotsset{height=3.8cm,width=3.6cm}
        \begin{tikzpicture}
          \begin{axis}[compat=newest,view = {0}{90},
              label style={font=\scriptsize}, xlabel={$x_1$}, ylabel={$x_2$},
              xmin=-3.5, xmax=3.5, ymin=-8, ymax=8,
            ]
            \addplot3 [gray,-stealth,
                quiver={u=\thisrow{u},v=\thisrow{v},scale arrows=0.02,},
                ] table {vdp/truth-vdp_vf.txt};
            \addplot [thick, blue] table {vdp/truth-vdp_gi0_traj.txt};
            \addplot [thick, green] table {vdp/truth-vdp_gi1_traj.txt};
            \draw [red,dashed] (-2.5,-4.5) rectangle (2.5,4.5);
          \end{axis}
        \end{tikzpicture}
        \subcaption{}\label{fig:vdp:truth}
    \end{minipage}
    \hspace{0.005\linewidth}
    \begin{minipage}[t]{0.17\linewidth}
        \centering\vspace*{0pt}
        \pgfplotsset{height=3.8cm,width=3.6cm}
        \begin{tikzpicture}
          \begin{axis}[compat=newest,view = {0}{90},
              label style={font=\scriptsize}, xlabel={$x_1$}, ylabel={$x_2$},
              xmin=-3.5, xmax=3.5, ymin=-8, ymax=8,
            ]
            \addplot3 [gray,-stealth,
                quiver={u=\thisrow{u},v=\thisrow{v},scale arrows=0.02,},
                ] table {vdp/d1=32-32_d2=64-64_d3=128_si=2.0_vf.txt};
            \addplot [thick, blue] table {vdp/d1=32-32_d2=64-64_d3=128_si=2.0_gi0_traj.txt};
            \addplot [thick, green] table {vdp/d1=32-32_d2=64-64_d3=128_si=2.0_gi1_traj.txt};
          \end{axis}
        \end{tikzpicture}
        \subcaption{}\label{fig:vdp:vf}
    \end{minipage}
    \hspace{0.005\linewidth}
    \begin{minipage}[t]{0.2\linewidth}
        \centering\vspace*{-1.6mm}
        \pgfplotsset{height=3.8cm,width=3.6cm}
        \begin{tikzpicture}
            \begin{axis}[compat=newest,view={0}{90},
                    label style={font=\scriptsize}, xlabel={$x_1$}, ylabel={$x_2$},
                    colorbar, colorbar style={width=2mm,ytick={1,2,3}},
                    xmin=-2.5, xmax=2.5, ymin=-5, ymax=5,
                ]
                \addplot3 [mesh/rows=50,mesh/num points=2500, contour gnuplot={number=20,output point meta=rawz,labels=false}] table {vdp/d1=32-32_d2=64-64_d3=128_si=2.0_lyap_Vvalue.txt};
                \addplot [black,dashed,thick] table {vdp/vdplc.txt};
            \end{axis}
        \end{tikzpicture}
        \subcaption{}\label{fig:vdp:vval}
    \end{minipage}
    \hspace{0.005\linewidth}
    \begin{minipage}[t]{0.2\linewidth}
        \centering\vspace*{-1.6mm}
        \pgfplotsset{height=3.8cm,width=3.6cm}
        \begin{tikzpicture}
            \begin{axis}[compat=newest,view={0}{90},
                    label style={font=\scriptsize}, xlabel={$x_1$}, ylabel={$x_2$},
                    colorbar, colorbar style={width=2mm,ytick={1,2,3}},
                    xmin=-2.5, xmax=2.5, ymin=-5, ymax=5,
                ]
                \addplot3 [mesh/rows=50,mesh/num points=2500, contour gnuplot={number=20,output point meta=rawz,labels=false}] table {vdp/d1=32-32_d2=4_d3=128_si=2.0_lyap_Vvalue.txt};
                \addplot [black,dashed,thick] table {vdp/vdplc.txt};
            \end{axis}
        \end{tikzpicture}
        \subcaption{}\label{fig:vdp:vval_wophi}
    \end{minipage}
    \hspace{0.005\linewidth}
    \begin{minipage}[t]{0.18\linewidth}
        \centering\vspace*{0pt}
        \pgfplotsset{height=3.8cm,width=3.6cm}
        \begin{tikzpicture}
            \begin{axis}[compat=newest,
                label style={font=\scriptsize}, xlabel={step}, ylabel={error},
                enlarge x limits=false,
                ytick={0,1},
                legend entries={vanilla,proposed},
                legend style={at={(0.02,0.71)}, anchor=south west, nodes={scale=0.5, transform shape}},
                ]
                \addplot [red,mark=x,mark repeat=2] table [x index=0, y index=1] {vdp/d1=32-32_d2=64-64_nosta_noinv_epi16_test_difabs_stat.txt};
                \addplot [green,mark=o,mark repeat=2] table [x index=0, y index=1] {vdp/d1=32-32_d2=64-64_d3=128_si=2.0_epi16_test_difabs_stat.txt};
            \end{axis}
        \end{tikzpicture}
        \subcaption{}\label{fig:vdp:error}
    \end{minipage}
    \caption{Results of Section~\ref{expt:vdp}. (a) True vector field of \eqref{eq:vdp} and two trajectories. Red dashed-line rectangle is the training data region. (b) Learned vector field and trajectories from it. (c) Learned $V(\bm{x})$. (d) Learned $V(\bm{x})$ without $\bm\phi$. (e) Prediction errors.}
    \label{fig:vdp}
\end{figure*}
\begin{figure*}
    \centering
    \begin{minipage}[t]{0.72\linewidth}
        \vspace*{0pt}\centering
        \setlength{\tabcolsep}{0pt}
        \begin{minipage}[t]{\linewidth}
            \vspace*{0pt}
            \begin{tabular}{cc}
                {\small \mylabeltext[\ref*{fig:flow}a]{(a)}{fig:flow:a}} &
                \raisebox{-.45\height}{
                \includegraphics[clip,trim={55pt 21pt 38pt 14pt},width=2cm]{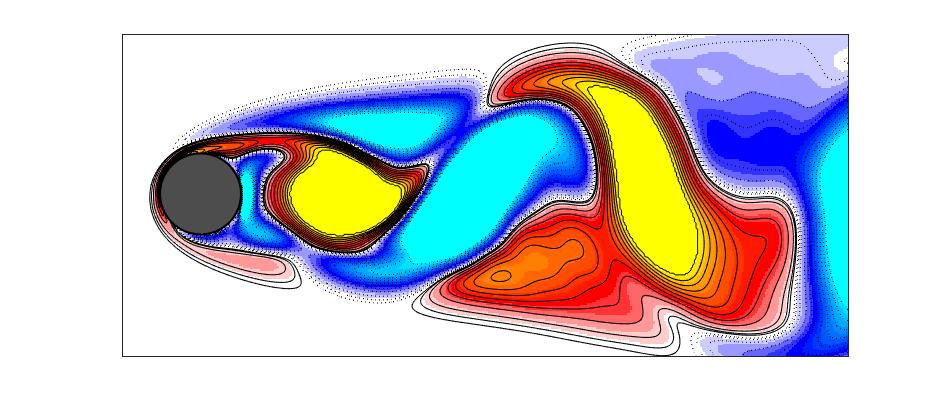}
                \includegraphics[clip,trim={55pt 21pt 38pt 14pt},width=2cm]{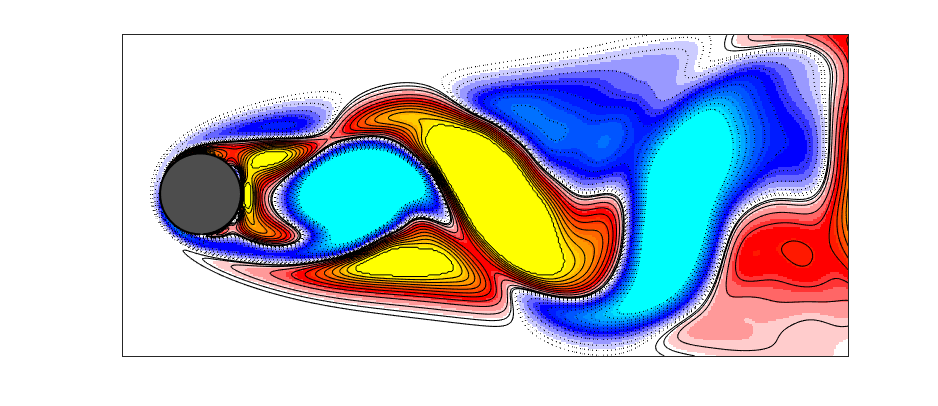}
                \includegraphics[clip,trim={55pt 21pt 38pt 14pt},width=2cm]{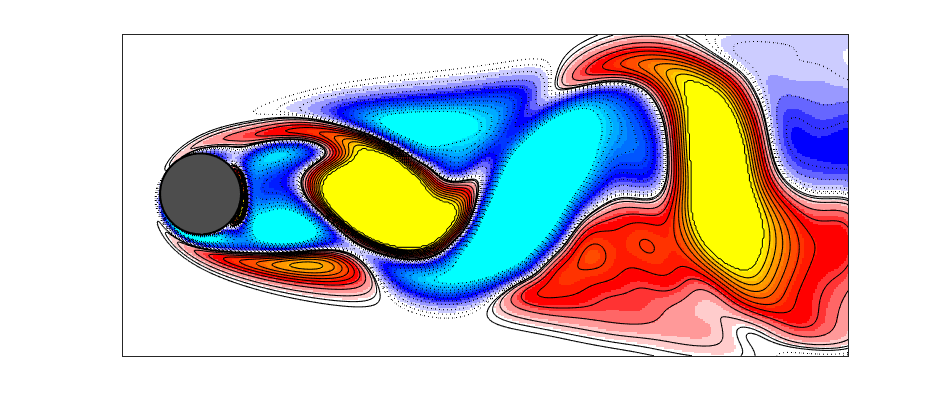}
                \includegraphics[clip,trim={55pt 21pt 38pt 14pt},width=2cm]{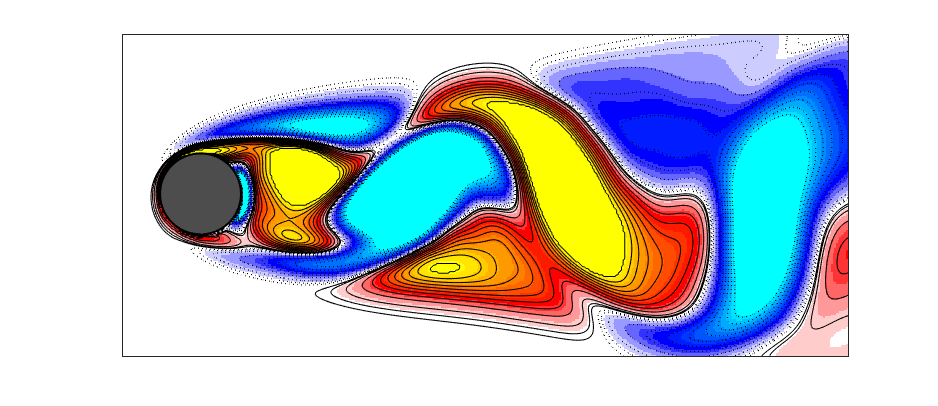}
                \includegraphics[clip,trim={55pt 21pt 38pt 14pt},width=2cm]{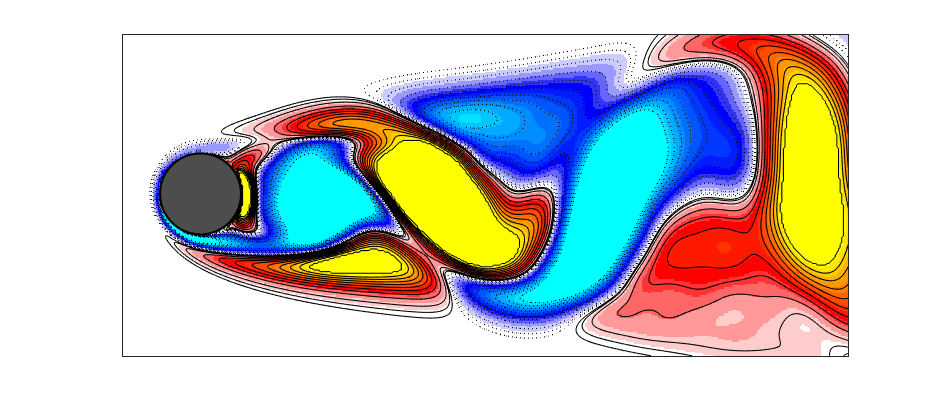}
                \includegraphics[clip,trim={55pt 21pt 38pt 14pt},width=2cm]{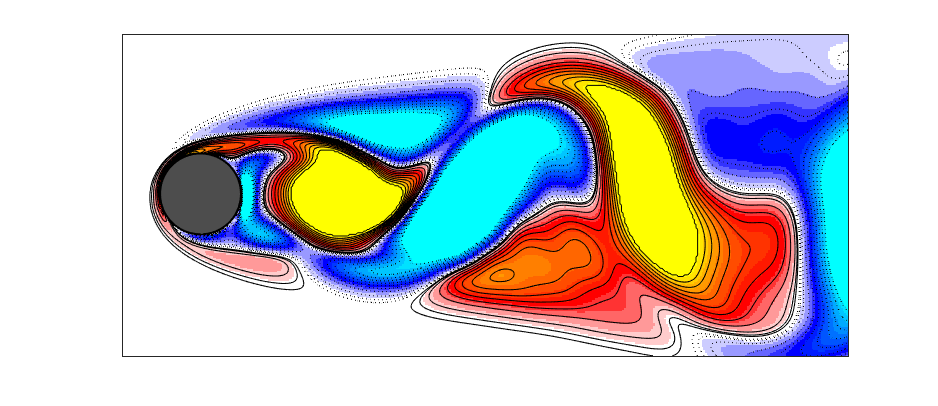}
                }
            \end{tabular}
        \end{minipage}\vspace*{1pt}
        \begin{minipage}[t]{\linewidth}
            \vspace*{0pt}
            \begin{tabular}{cc}
                {\small \mylabeltext[\ref*{fig:flow}b]{(b)}{fig:flow:b}} &
                \raisebox{-.45\height}{
                \includegraphics[clip,trim={55pt 21pt 38pt 14pt},width=2cm]{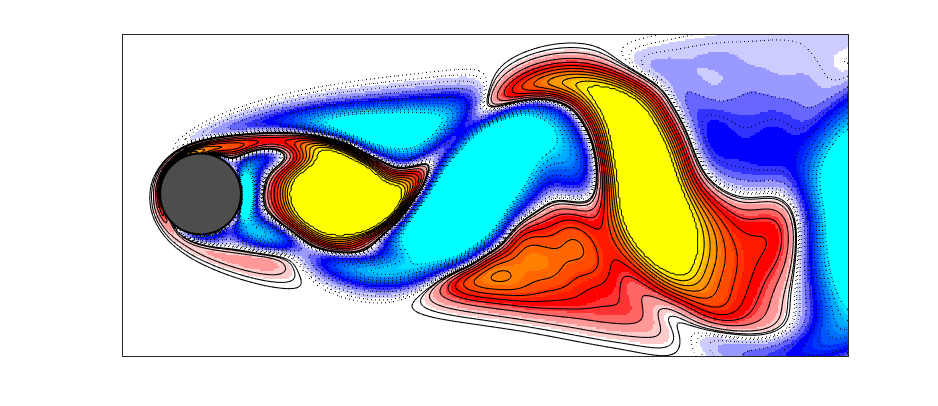}
                \includegraphics[clip,trim={55pt 21pt 38pt 14pt},width=2cm]{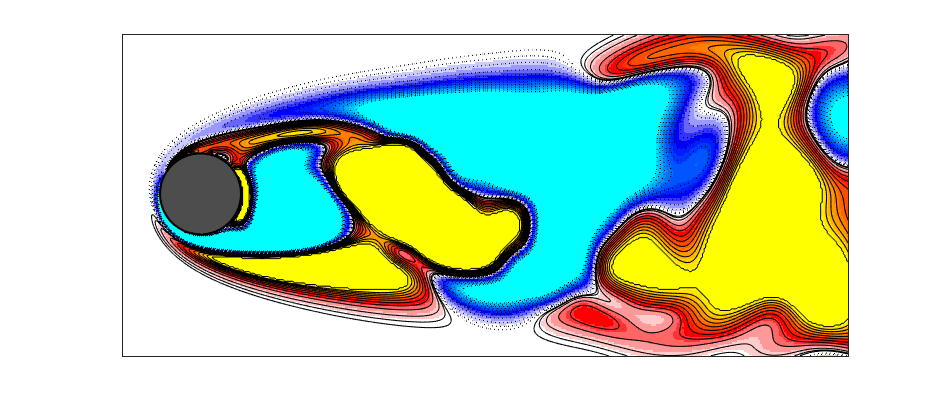}
                \includegraphics[clip,trim={55pt 21pt 38pt 14pt},width=2cm]{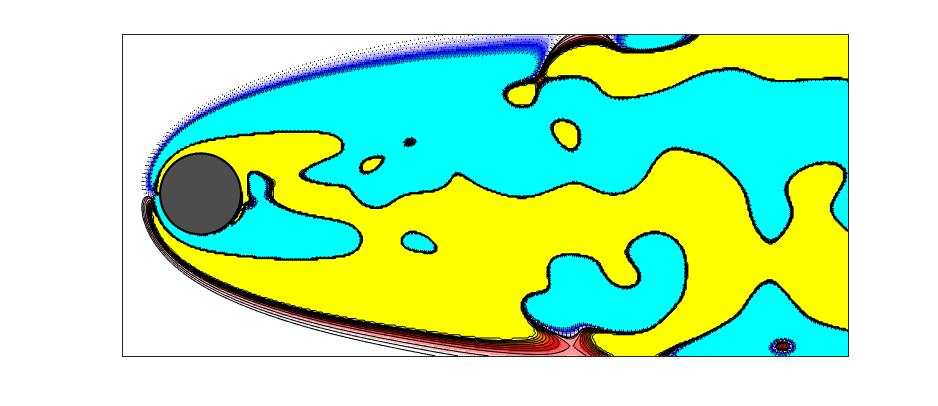}
                \includegraphics[clip,trim={55pt 21pt 38pt 14pt},width=2cm]{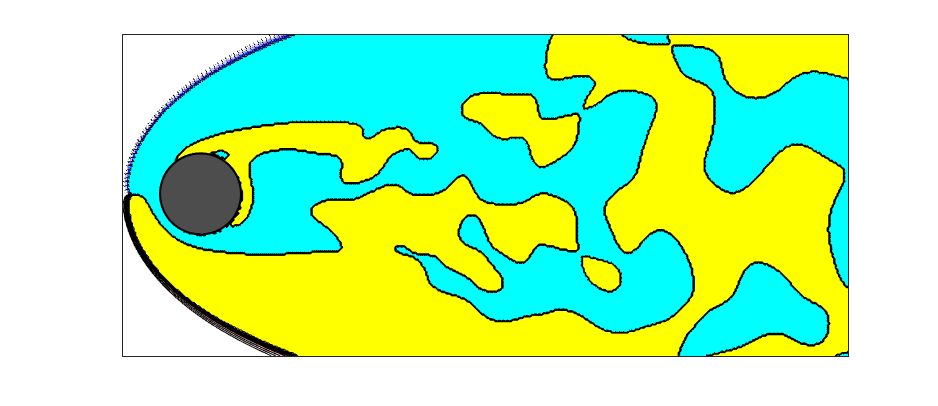}
                \includegraphics[clip,trim={55pt 21pt 38pt 14pt},width=2cm]{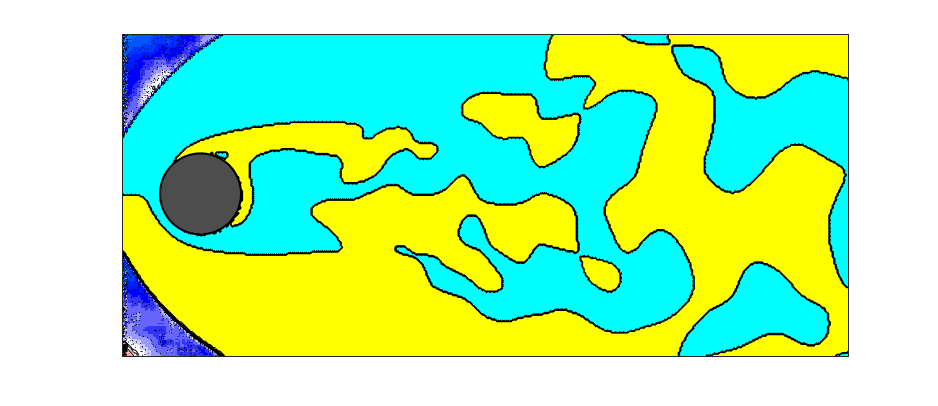}
                \includegraphics[clip,trim={55pt 21pt 38pt 14pt},width=2cm]{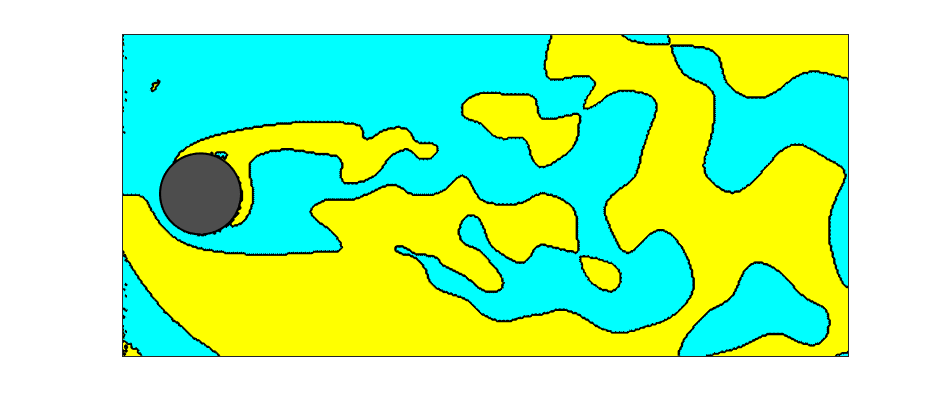}
                }
            \end{tabular}
        \end{minipage}\vspace*{1pt}
        \begin{minipage}[t]{\linewidth}
            \vspace*{0pt}
            \begin{tabular}{cc}
                {\small \mylabeltext[\ref*{fig:flow}c]{(c)}{fig:flow:c}} &
                \raisebox{-.45\height}{
                \includegraphics[clip,trim={55pt 21pt 38pt 14pt},width=2cm]{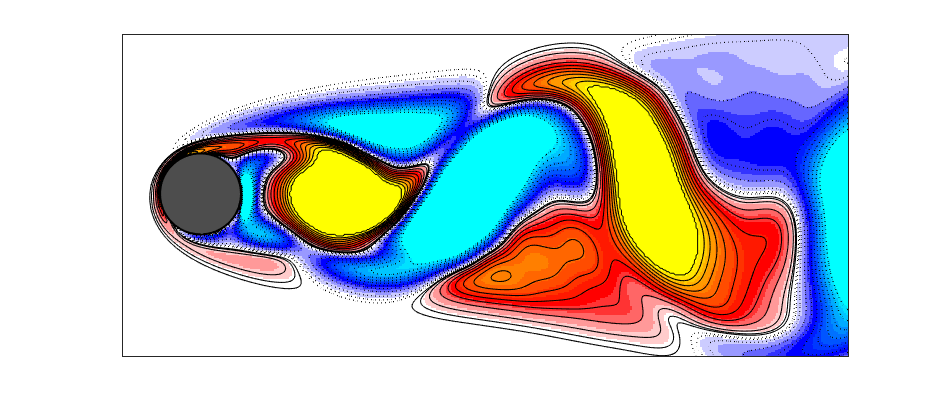}
                \includegraphics[clip,trim={55pt 21pt 38pt 14pt},width=2cm]{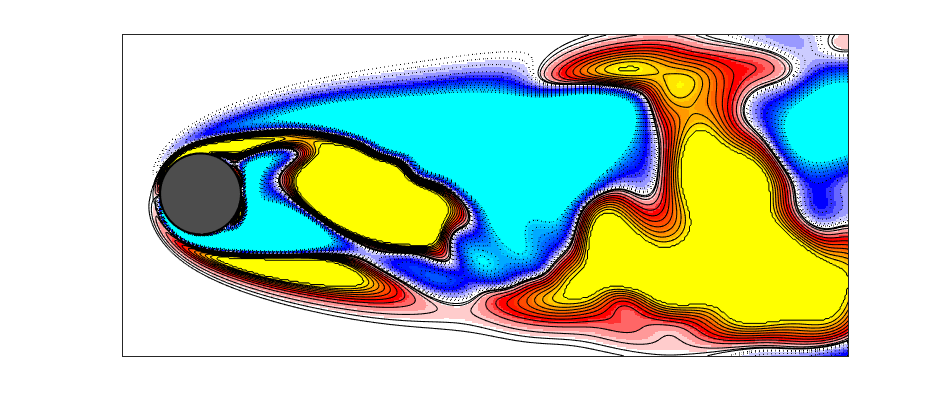}
                \includegraphics[clip,trim={55pt 21pt 38pt 14pt},width=2cm]{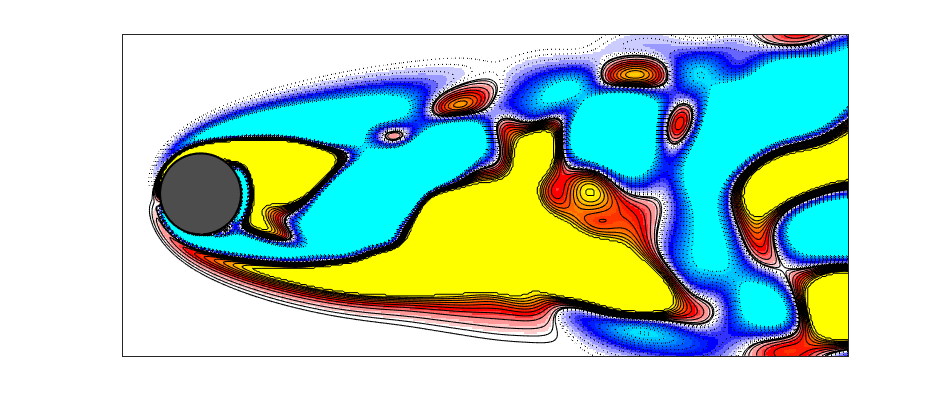}
                \includegraphics[clip,trim={55pt 21pt 38pt 14pt},width=2cm]{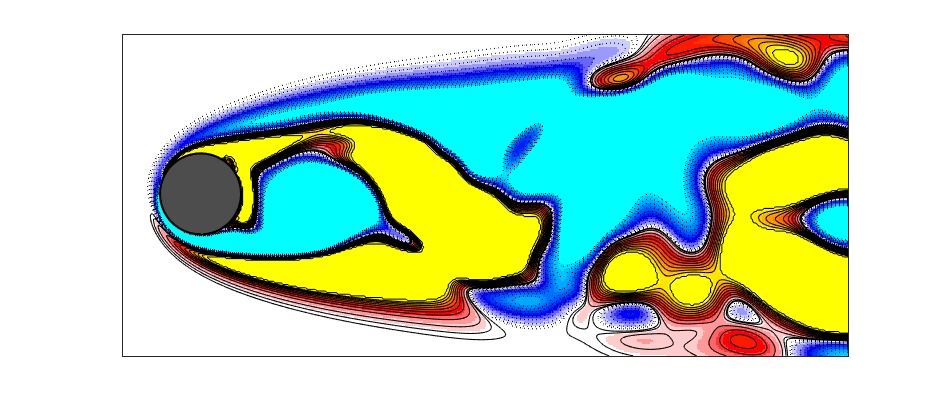}
                \includegraphics[clip,trim={55pt 21pt 38pt 14pt},width=2cm]{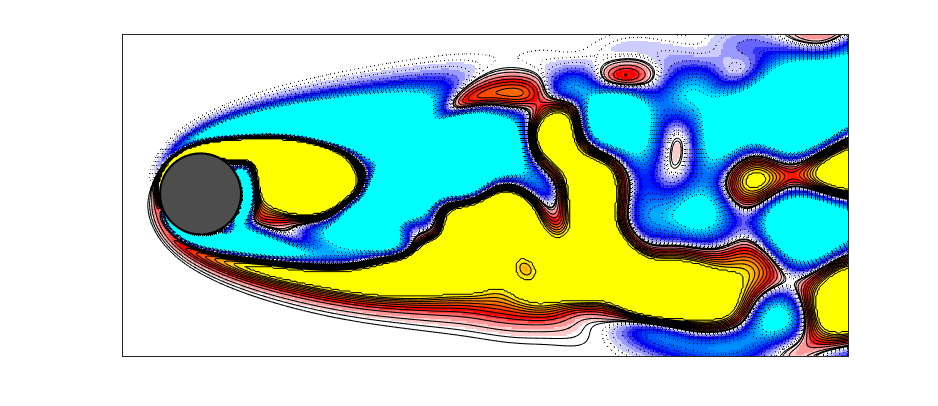}
                \includegraphics[clip,trim={55pt 21pt 38pt 14pt},width=2cm]{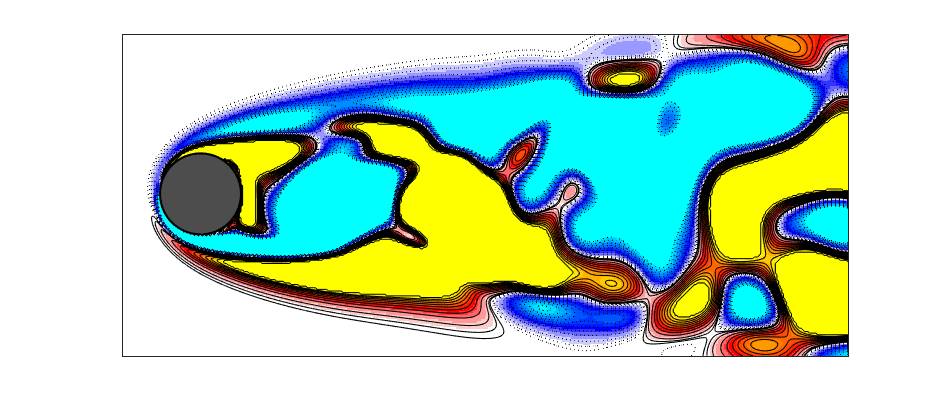}
                }
            \end{tabular}
        \end{minipage}\vspace*{1pt}
        \begin{minipage}[t]{\linewidth}
            \vspace*{0pt}
            \begin{tabular}{cc}
                {\small \mylabeltext[\ref*{fig:flow}d]{(d)}{fig:flow:d}} &
                \raisebox{-.45\height}{
                \includegraphics[clip,trim={55pt 21pt 38pt 14pt},width=2cm]{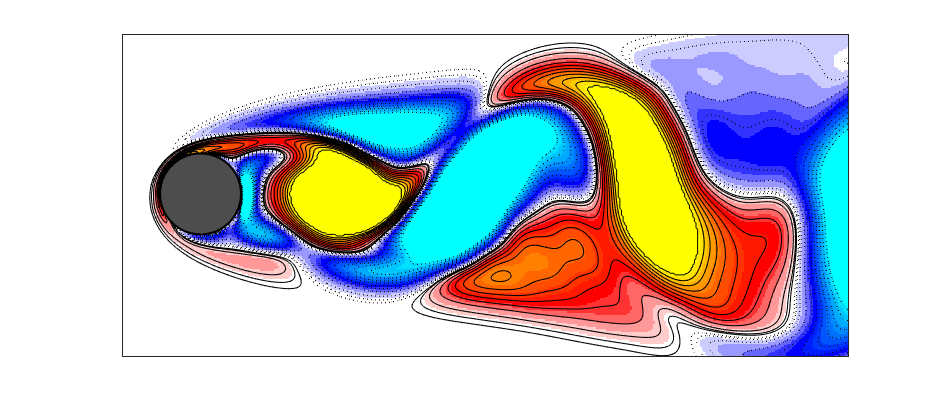}
                \includegraphics[clip,trim={55pt 21pt 38pt 14pt},width=2cm]{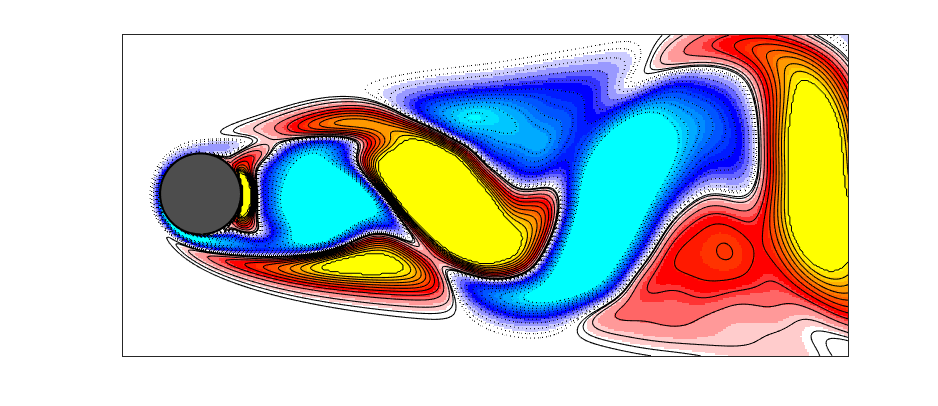}
                \includegraphics[clip,trim={55pt 21pt 38pt 14pt},width=2cm]{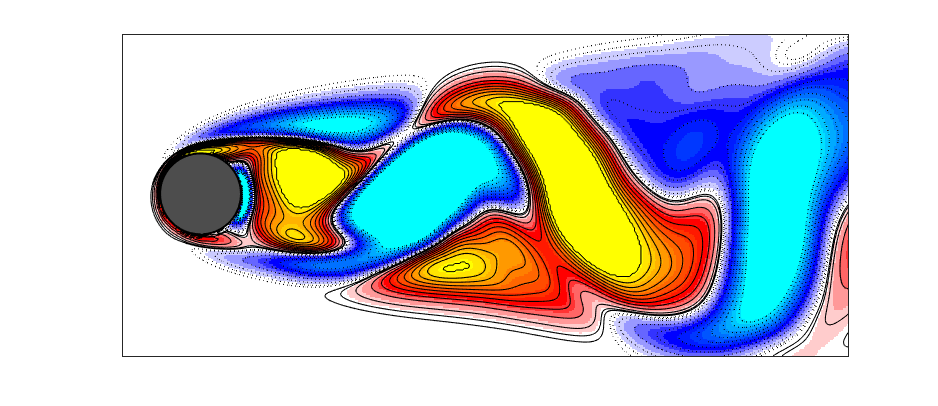}
                \includegraphics[clip,trim={55pt 21pt 38pt 14pt},width=2cm]{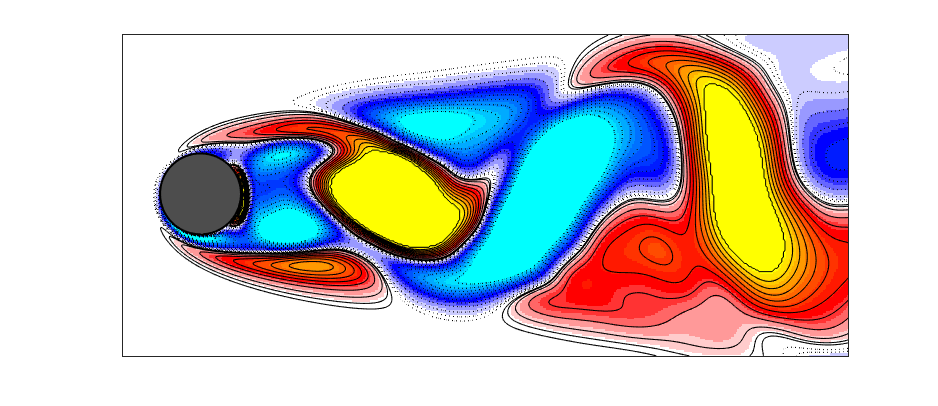}
                \includegraphics[clip,trim={55pt 21pt 38pt 14pt},width=2cm]{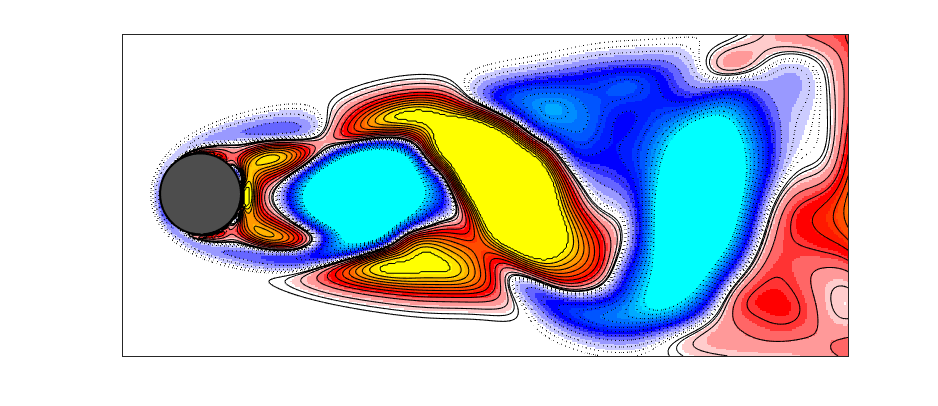}
                \includegraphics[clip,trim={55pt 21pt 38pt 14pt},width=2cm]{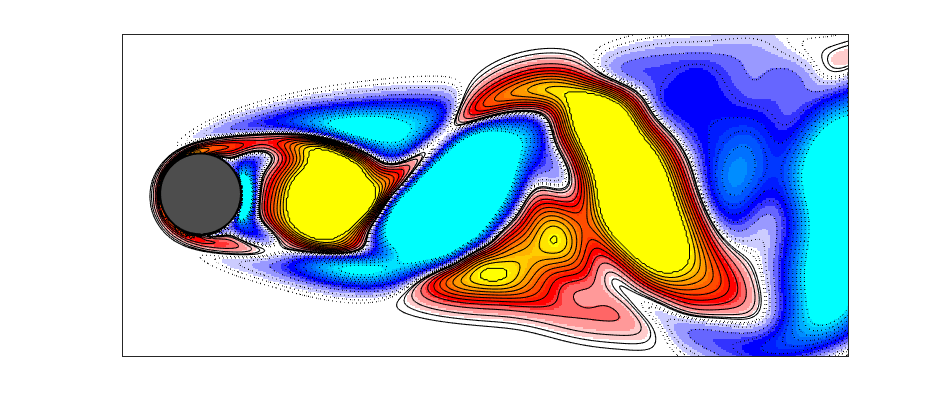}
                }
            \end{tabular}
        \end{minipage}\vspace*{2pt}
        \begin{minipage}[t]{\linewidth}
            \vspace*{0pt}
            \newcolumntype{x}{>{\centering\arraybackslash}p{2.1cm}}
            {\small\begin{tabular}{p{4mm}xxxxxx}
                 & $t=0$ & $t=40$ & $t=80$ & $t=120$ & $t=160$ & $t=200$
            \end{tabular}}
        \end{minipage}
    \end{minipage}
    \hspace{0.02\linewidth}
    \begin{minipage}[t]{0.24\linewidth}
        \vspace*{0pt}\centering
        \caption{Long-term predictions of on-attractor fluid flow given off-attractor training data (Section~\ref{expt:fluid}). Red\&yellow and blue\&cyan denote positive and negative values of vorticity, respectively. (a) Ground truth. (b) The vanilla model. (c) The stable equilibrium model \citep{manekLearningStableDeep2019}. (d) The proposed model.}
        \label{fig:flow}
    \end{minipage}
\end{figure*}

\subsection{Learning Vector Field of Nonlinear Oscillator}
\label{expt:vdp}

The Van der Pol oscillator:
\begin{equation}\label{eq:vdp}
    \dot{x}_1 = x_2,\quad
    \dot{x}_2 = \mu (1-x_1^2) x_2 - x_1
\end{equation}
is well known as a basis for modeling many physical and biological phenomena.
It has a stable limit cycle, whose exact shape cannot be described analytically.

As training data (Figure~\ref{fig:vdp:truth}), we used the values of $\bm{x}$ and $\dot{\bm{x}}$ sampled from an even grid on area $[-2.5,2.5] \times [-4.5,4.5]$ (i.e., the red dashed-line rectangle in Figure~\ref{fig:vdp:truth}) with $\mu=2$.
For the proposed model, we set $\tilde{\invset}$ to be a circle defined as in \eqref{eq:invset:surf}, expecting $\bm\phi$ would be learned so that it transforms a circle into the limit cycle of the system.

In Figure~\ref{fig:vdp:vf}, we show the learned vector field and two trajectories generated from it.
They successfully resemble the truth (in Figure~\ref{fig:vdp:truth}), even outside the area of the training data.
In Figure~\ref{fig:vdp:vval}, we depict the values of learned $V(\bm{x})$, wherein we can observe $V(\bm{x})$ decreases toward the limit cycle of the system (the dashed orbit).
In Figure~\ref{fig:vdp:vval_wophi}, for comparison, we show learned $V(\bm{x})$ \emph{without} a learnable feature transform $\bm\phi$; not surprisingly, it fails to capture the shape of the limit cycle.
In Figure~\ref{fig:vdp:error}, we show an example of long-term prediction errors against prediction steps (a single step corresponds to the $\Delta t$).
The model with the proposed stability guarantee achieves significantly lower long-term prediction errors.


\subsection{Application: Fluid Flow Prediction}
\label{expt:fluid}

We apply the proposed stable dynamics model to an application of fluid flow prediction.
The target flow is so-called cylinder wake (see Figure~\ref{fig:flow:a}); a cylinder-like object is located in a 2D field, fluids come from one side uniformly, and there occurs a series of vortices past the object in certain conditions.
This is a limit cycle known as the K\'arm\'an's vortex street.
Before the flow reaches the limit cycle, it typically starts from an unstable equilibrium, and then the vortices begin to grow gradually.
This stage is called off-attractor.
Cylinder wake has been studied as one of standard problems of fluid dynamics and also appears as a testbed of forecasting methods even recently \citep[see, e.g.,][]{dmdbook}.

As training data, we generated such flow using the immersed boundary projection method \citep{Taira07,Colonius08} and used the part from near the equilibrium to a time point \emph{before} the limit cycle is completely observed; hence the training data were off-attractor.
The data comprised the observations of the vorticity in the field of size $199 \times 449$.
As preprocessing, we reduced the dimensionality of data from $89351$ to $26$ by PCA, which lost only $0.1$\% of the energy.
We contaminated the data with Gaussian noise.
We estimated $\dot{\bm{x}}$ by $(\bm{x}_{t+\Delta t}-\bm{x}_t)/\Delta t$ and learned the proposed dynamics model with a cycle along the first two axes of $\mathcal{Z}$ as $\tilde{\invset}$ (i.e., $C_{\tilde{\invset}}(\bm{z})=z_1^2+z_2^2-r^2$).

In Figure~\ref{fig:flow}, we show the results of long-term prediction starting at a time point where the flow is almost on the limit cycle (i.e., on-attractor).
The two baselines (in Figures~\ref{fig:flow:b} and \ref{fig:flow:c}) fail to replicate the true limit cycle (in Figure~\ref{fig:flow:a}).
In contrast, the long-term prediction by the proposed method (in Figure~\ref{fig:flow:d}) shows a plausible oscillating pattern, though the oscillation phase is slightly different from the truth.
It is worth noting that with the proposed stable dynamics model, we were able to predict the on-attractor oscillating patterns only from off-attractor training data.


\section{Conclusion}
\label{concl}

We proposed a dynamics model with the provable existence of a stable invariant set.
It can handle the stability of general types of invariant sets, for example, limit cycles and line attractors.
Future directions of research include the treatment of random dynamical systems as the current method is limited to deterministic dynamics.
Consideration of the input-to-state stability of controlled systems is also an interesting problem.


\bibliography{main,book,additional}

\appendix


\section{Proof of Proposition 1}

\fbox{\begin{minipage}{0.981\linewidth}
\begin{propositionappendix}\label{_prop:inv}
    For a dynamical system $\dot{\bm{z}}=\bm{F}(\bm{z})$ with some $\bm{F}:\mathcal{Z}\to\mathcal{Z}$,
    \begin{enumerate}[topsep=0pt,itemsep=0pt,leftmargin=2em,label=(\alph*)]
        \item\label{_item:prop:inv:a} If $C_{\tilde{\invset}}(\bm{z}) = 0 \, \Rightarrow \, \surfcond{\bm{F}}{\bm{z}} > 0$, then $\tilde{\invset}^\text{vol}$ in \eqref{eq:invset:vol} is a positively invariant set.
        \item\label{_item:prop:inv:b} If $C_{\tilde{\invset}}(\bm{z}) = 0 \, \Rightarrow \, \surfcond{\bm{F}}{\bm{z}} = 0$, then $\tilde{\invset}^\text{surf}$ in \eqref{eq:invset:surf} is a positively invariant set.
    \end{enumerate}
\end{propositionappendix}
\end{minipage}}

\begin{proof}
    Let $\bm{z}(t)$ be a trajectory of the dynamical system, $\dot{\bm{z}}=\bm{F}(\bm{z})$.
    Let $c:\mathbb{R}\to\mathbb{R}$ be a function such that $c(\tau)=C_{\tilde{\invset}}(\bm{z}(\tau))$.
    
    First, let us consider case \ref{_item:prop:inv:a}.
    For a proof by contradiction, assume the negation of \ref{_item:prop:inv:a}, that is, $C_{\tilde{\invset}}(\bm{z}) = 0 \, \Rightarrow \, \surfcond{\bm{F}}{\bm{z}} > 0$ \emph{and} $\tilde{\invset}=\tilde{\invset}^\text{vol}$ is \emph{not} a positively invariant set (i.e., $\bm{z}(t) \in \tilde{\invset}$ and $\bm{z}(s) \notin \tilde{\invset}$ for some $t \leq s$).
    Then, from the definition of $\tilde{\invset}^\text{vol}$ in \eqref{eq:invset:vol}, we have $c(t) \geq 0$ and $c(s) < 0$.
    By the continuity of $c$, there is at least one point $r \in [t,s]$ where $c(r)=0$ and $\dot{c}(r) \leq 0$.
    At this point, $C_{\tilde{\invset}}(\bm{z}(r))=0$ and $\nabla C_{\tilde{\invset}}(\bm{z}(r))^\tr \bm{F}(\bm{z}(r)) = \dot{c}(r) \leq 0$, which is a contradiction to what we assumed.
    Therefore, \ref{_item:prop:inv:a} holds.
    
    Second, let us see \ref{_item:prop:inv:b}.
    Analogously to the above case, assume $C_{\tilde{\invset}}(\bm{z}) = 0 \, \Rightarrow \, \surfcond{\bm{F}}{\bm{z}} = 0$ \emph{and} $\tilde{\invset}=\tilde{\invset}^\text{surf}$ is \emph{not} a positively invariant set.
    Then, from the definition of $\tilde{\invset}^\text{surf}$ in \eqref{eq:invset:surf}, we have $c(t) = 0$ and $c(s) \neq 0$.
    Hence, we have $c(r)=0$ and $\dot{c}(r) \neq 0$ at some point $r \in [t,s]$, which is a contradiction.
    Therefore, \ref{_item:prop:inv:b} holds.
\end{proof}


\section{Proof of Proposition 2}

\fbox{\begin{minipage}{0.981\linewidth}
\begin{propositionappendix}\label{_prop:main}
    Let $\tilde\invset^\text{vol}$ (or $\tilde\invset^\text{surf}$) be a subset of $\mathcal{Z} \subseteq \mathbb{R}^d$ defined in \eqref{eq:invset:vol} (or \eqref{eq:invset:surf}).
    Let $\tilde{\bm{f}}:\mathbb{R}^d\to\mathbb{R}^d$ be the function in \eqref{eq:mod2}.
    Then, for a dynamical system $\dot{\bm{z}}=\tilde{\bm{f}}(\bm{x})$, $\tilde\invset^\text{vol}$ (or $\tilde\invset^\text{surf}$) is a positively invariant set and is asymptotically stable.
\end{propositionappendix}
\end{minipage}}

\begin{proof}
    Let us consider the case of $\tilde\invset^\text{vol}$ (the discussion holds analogously for $\tilde\invset^\text{surf}$).
    Recall that from the definition, $C_{\tilde{\invset}}(\bm{z})=0$ implies $\bm{z} \in \tilde{\invset}$.
    Hence, from \eqref{eq:mod2}, if $C_{\tilde{\invset}}(\bm{z})=0$, then $\nabla C_{\tilde{\invset}}(\bm{z})^\tr \tilde{\bm{f}}(\bm{z}) = \xi(\bm{z}) > 0$, which proves the invariance of $\tilde{\invset}$ (see Proposition~\ref{prop:inv}).
    As for stability, we should show
    \begin{equation}\label{eq:prop2:key}\begin{aligned}
        \dot{V}(\bm{z}) &= 0, \quad\text{if and only if}\quad \bm{z} \in \tilde{\invset},\\ 
        \dot{V}(\bm{z}) &< 0, \quad\text{otherwise}.
    \end{aligned}\end{equation}
    Suppose $\bm{z} \in \tilde{\invset}$.
    We have $V(\bm{z})=0$ for every $\bm{z}\in\tilde{\invset}$ from the construction, and the orbits of $\tilde{\bm{f}}$ stay in $\tilde{\invset}$ because $\tilde{\invset}$ is a positively invariant set. Hence, $\dot{V}(\bm{z})=0$ for $\bm{z} \in \tilde{\invset}$.
    On the other hand, suppose $\bm{z} \notin \tilde{\invset}$.
    Then, we have
    \begin{equation*}\begin{aligned}
        &
        \nabla V(\bm{z})^\tr \tilde{\bm{f}}(\bm{z}) + \alpha V(\bm{z})
        \\
        &\quad=
        \nabla V(\bm{z})^\tr \bm{g}(\bm{z}) + \alpha V(\bm{z})
        \\
        &\quad=
        \beta(\bm{z}) - u \big( \beta(\bm{z}) \big) \big( \beta(\bm{z}) + \eta(\bm{z}) \big).
    \end{aligned}\end{equation*}
    First, suppose $\beta(\bm{z}) \geq 0$.
    Then $u(\beta(\bm{z}))=1$, and thus $\nabla V(\bm{z})^\tr \tilde{\bm{f}}(\bm{z}) + \alpha V(\bm{z}) = - \eta(\bm{z}) \leq 0$.
    Second, suppose $\beta(\bm{z}) < 0$. Then  $u(\beta(\bm{z}))=0$, and thus $\nabla V(\bm{z})^\tr \tilde{\bm{f}}(\bm{z}) + \alpha V(\bm{z}) = \beta(\bm{z}) < 0$.
    As $V(\bm{z})>0$ at $\bm{z} \notin \tilde{\invset}$ from the construction of $V$, in either of the cases above, we have $\nabla V(\bm{z})^\tr \tilde{\bm{f}}(\bm{z}) \leq -\alpha V(\bm{z}) < 0$.
    Hence, $\dot{V}(\bm{z}) < 0$ for all $\bm{z}\notin\tilde{\invset}$.
    This proves \eqref{eq:prop2:key}, from which we can say that $\tilde{\invset}$ is the largest subset of the state space such that $\dot{V}(\bm{z})=0$.
    Therefore, from Theorem~\ref{thm:lasalle}, $\tilde{\invset}$ is asymptotically stable.
\end{proof}


\section{Additional Related Work}

We introduce related studies that could not be in the main text due to the space limitation.
Note that \emph{all} the studies below also consider stability (if any) of equilibrium only.

We should mention the work of \citet{lawrence_almost_2020}.
They proposed a dynamics model for discrete-time stochastic systems with almost sure stability.
They proposed a way to use non-convex neural Lyapunov function $V$, for which they ensured the stability condition on $V$ (i.e., $V$ should decrease along time) via solution of an optimization problem, instead of the projection as in \citet{manekLearningStableDeep2019} and ours.
Consequently, training their models depends on the implicit function theorem to backpropagate through the optimization problem.

There are other several attempts (that could not mentioned in the main text) to learn stable dynamics models.
\citet{umlauft_learning_2017} also considered stability of GP-based state space models.
\citet{yu_onsagernet_2020} proposed a method to learn neural stable dynamics model with monotonicity property.
\citet{mehrjou_learning_2020} proposed a method to learn a dynamics model, a Lyapunov function, and the level sets of the learned Lyapunov function simultaneously.
\citet{lemme_neural_2014} proposed to learn vector fields with stable equilibrium by imposing constraints on learnable quadratic Lyapunov function via sampling most condition-violating points during optimization and adding them to constraints of the optimization.
They also considered positive invariance of a neighborhood of (to-be) stable equilibrium as constraints of the optimization.

In more control-like contexts, \citet{xiao_learning_2020} proposed a method to build a GP-based control Lyapunov function to stabilize GP-based dynamics models.
In reinforcement learning context, \citet{berkenkamp_safe_2017} considered safety issues by computing regions of attractions under a specific policy.
In an imitation learning setting, \citet{khansari-zadeh_learning_2014} parametrized a control Lyapunov function as a weighted sum of asymmetric quadratic functions and learned its parameters from demonstrations.

While not directly connected to our problem of learning stable dynamics models, the stability and related properties of deep neural networks have been of great interest to understand and improve their performance in supervised and unsupervised tasks.
For example, \citet{haber_stable_2018} discussed the stability of deep neural networks to understand exploding / vanishing gradient problems.
\citet{miller_stable_2019} studied stable recurrent neural networks and showed superior performance of stable ones over unstable counterparts for tasks such as language modeling.
\citet{bonassi_lstm_2020} proposed a method to ensure the input-to-state stability property of LSTM networks, which is useful for controlling the network's reachable set.
Furthermore, \citet{hanson_universal_2020} showed that recurrent neural networks are a universal approximator of infinite trajectories from stable dynamical systems.

In the context similar to learning physically-meaningful models, \citet{chen_symplectic_2020} proposed Hamiltonian neural networks equipped with symplectic integrators.
Also, \citet{chang_reversible_2018} discussed reversible neural networks.


\section{Implementation Details}

In the experiments, we used the proposed dynamics model and the baseline models implemented using neural networks as their components.
They were implemented mainly using PyTorch, and the codes are also attached as supplementary materials.

\paragraph{Implementation of $\bm\phi$}
For the invertible feature transform $\bm\phi$, if any (as it was not used in some experiments), we used the augmented neural ODEs (ANODE) \citep{dupontAugmentedNeuralODEs2019}.
That is, $\bm{z}=\bm\phi(\bm{x})$ is given as
\begin{equation}\label{_eq:anode1}\begin{gathered}
    \begin{bmatrix}
        \underbrace{\bm{z}^\tr}_{1 \times d} & \underbrace{\bm{a}^\tr}_{1 \times d_\text{aug}}
    \end{bmatrix}^\tr
    = \int_0^1
    \bm\psi \left( \begin{bmatrix} \bm{u}_z(\tau)^\tr & \bm{u}_a(\tau)^\tr \end{bmatrix}^\tr \right) \mathrm{d}\tau,
    \\
    \text{with}\quad
    \bm{u}_z(0)=\bm{x}
    \quad\text{and}\quad
    \bm{u}_a(0)=\bm{0}.
\end{gathered}\end{equation}
Here, $d_\text{aug}$ denotes the dimensionality of the augmentation variable $\bm{a}$ and $\bm{u}_a(\tau)$.
It can represent non-homeomorphic functions with $d_\text{aug}>0$ \citep{papamakarios_normalizing_2019}.
Its inverse, $\bm{x}=\bm\phi^{-1}(\bm{z})$, is given as
\begin{equation}\label{_eq:anode2}\begin{gathered}
    \bm{x}
    = \left[ \int_1^0
    \bm\psi \left( \begin{bmatrix} \bm{u}_z(\tau)^\tr & \bm{u}_a(\tau)^\tr \end{bmatrix}^\tr \right) \mathrm{d}\tau \right]_{1:d},
    \\
    \text{with}\quad
    \bm{u}_z(1)=\bm{z}
    \quad\text{and}\quad
    \bm{u}_a(1)=\bm{a},
\end{gathered}\end{equation}
where $[\cdot]_{1:d}$ means the first $d$ elements of a vector. 
As the function $\bm\psi:\mathbb{R}^{d+d_\text{aug}} \to \mathbb{R}^{d+d_\text{aug}}$, we used feed-forward neural networks with fully-connected layers.
We used the exponential linear unit (ELU) with PyTorch's default parameter as activation function.
We selected the number of hidden layers and the number of the units of those layers in accordance with the loss values on validation data.

\paragraph{Implementation of $\bm{h}$}
For the base dynamics model $\bm{h}$, we used feed-forward neural networks with fully-connected layers and ELU as activation function.
We selected the number of hidden layers and the number of the units of those layers in accordance with the loss values on validation data.
We applied the batch normalization technique except before an output layer.

\paragraph{Implementation of $V$ ($q$ and $\eta$)}
For implementing $V$, we need to implement $q$ and $\eta$.
For $q$, we used the input-convex neural networks \citep{amosInputConvexNeural2017} with ELU as activation function.
That is,
\begin{equation}
    q(\bm{z}) = \big( q_K \circ \cdots \circ q_1 \big) (\bm{z}),
\end{equation}
where, letting $\bm{y}_1=\bm{z}$, $\bm{y}_2=q_1(\bm{y}_1)$ and so on,
\begin{equation}
    q_k(\bm{y}_k) =
    \begin{cases}
        \operatorname{ELU} \left( \bm{A}_k \bm{y}_k \right), & k=1, \\
        \operatorname{ELU} \left( \bm{A}_k \bm{y}_k + \operatorname{softplus}(\bm{B}_k) \bm{z} \right), & 1<k<K, \\
        \bm{A}_k \bm{y}_k + \operatorname{softplus}(\bm{B}_k) \bm{z}, & k=K,
    \end{cases}
\end{equation}
with parameters $\bm{A}_k \in \mathbb{R}^{\dim(\bm{y}_{k+1}) \times \dim(\bm{y}_k)}$ and $\bm{B}_k \in \mathbb{R}^{\dim(\bm{y}_{k+1}) \times d}$.
Note $\dim(\bm{y}_{K+1})=\dim(q(\bm{z}))=1$ as $q$ must be a scalar-valued function.
We selected the number of hidden layers and the number of the units of those layers in accordance with the loss values on validation data.
For $\eta$, we used neural networks with fully-connected layers and ELU as activation function, with output value being clipped to be nonnegative.

\paragraph{Other Parameters}
We set $\alpha=0.01$ and $\varepsilon=0.1$ in every experiment.


\section{Experiment Details}

The experiments were performed using the Intel Xeon Gold 6148 processors with DDR4 2400 MHz 256 GB RAM.
In the following, we present the configurations in each experiment.
We denote the configuration of a feed-forward neural network by, for example, ``64-32'' when the network comprises an input layer, two hidden layers, and an output layer, and the numbers of the units of the two hidden layers are 64 and 32, respectively.
Note that the number of units of the input and output layers is obvious from data properties and the model architecture.
Below we detail the configurations of each experiment.


\subsection{Simple Example of Limit Cycle}

\paragraph{Data}
In generating data from the system with a limit cycle in Section~\ref{expt:simple}, we used MATLAB's \texttt{ode45} function with $\Delta t = 0.075$.
As training data, we generated four trajectories of length $20$ with initial conditions: $x_1,x_2=-2,0.5$; $x_1,x_2=2,0.5$; $x_1,x_2=-0.3,-0.3$; and $x_1,x_2=0.3,0.3$.
As validation data, we generated four trajectories of length $20$ with initial conditions: $x_1,x_2=-1.5,0$; $x_1,x_2=1.5,0$; $x_1,x_2=-0.5,-0.5$; and $x_1,x_2=0.5,0.5$.
We used the value of $\bm{x}$ and $\dot{\bm{x}}$ on these trajectories for training and validation.
As test data, we generated $20$ trajectories of length $50$ with initial conditions drawn randomly from the uniform distribution over the range $x_1,x_2 \in [-1.5,1.5] \times [-0.5,0.5]$.

\paragraph{Model and Hyperparameter}
In this experiment, we did not use $\bm\phi$, that is, we set $\bm\phi(\bm{x})=\bm{x}$.
We fixed the configuration of the network of $\bm{h}$ to be 64-64.
We selected the architecture of the network of $q$ by validation loss.
The search range was: 8, \textcolor{red}{16}, 32, 64, 128, 8-8, 16-16, 32-32, 64-64, and 128-128 (\textcolor{red}{red}: finally selected in our case).

\paragraph{Optimization}
We used the Adam optimizer \citep{Kingma15} with learning rate $0.0001$ and full-batch updates.
We set the decay rate parameter to be $0.00001$.
For the other parameters, we used the default values of PyTorch 1.4.0.
We adopted the early stopping strategy by watching the loss value on the validation data.

\paragraph{Computational Speed}
For a full-batch update of gradient descent (i.e., single epoch), the proposed method required about $20$ ms in average, whereas the vanilla method required about $6$ ms.


\subsection{Simple Example of Line Attractor}

\paragraph{Data}
In generating data from the system with a line attractor in Section~\ref{expt:simple}, we used MATLAB's \texttt{ode45} function with $\Delta t = 0.05$.
As training data, we generated $16$ trajectories of length $80$ with initial conditions on the points of a $4 \times 4$ even grid in area $[-2,2] \times [-2,2]$.
As validation data, we generated 16 trajectories of length $80$ with initial conditions on the points of a $4 \times 4$ even grid in area $[-1.5,1.5] \times [-2,2]$.
We used the value of $\bm{x}$ and $\dot{\bm{x}}$ on these trajectories for training and validation.
We made no test data as the main purpose of this experiment was to watch the values of learned $V(\bm{x})$.

\paragraph{Model and Hyperparameter}
In this experiment, we did not use $\bm\phi$, that is, we set $\bm\phi(\bm{x})=\bm{x}$.
We fixed the configuration of the network of $\bm{h}$ to be 64-64.
We selected the architecture of the network of $q$ by validation loss.
The search range was: \textcolor{red}{16} and 64.

\paragraph{Optimization}
We used the Adam optimizer with learning rate $0.001$ and full-batch updates.
We set the decay rate parameter to be $0.00001$.
We adopted the early stopping strategy with validation loss.

\paragraph{Computational Speed}
For a full-batch update of gradient descent (i.e., a single epoch), the proposed method required about $50$ ms in average, whereas the vanilla method required about $40$ ms.


\subsection{Learning vector field of nonlinear oscillator}

\paragraph{Data}
As training data, we computed the values of $\bm{x}$ and $\dot{\bm{x}}$ following \eqref{eq:vdp} on the points of a $20 \times 20$ even grid in area $[-2.5,2.5] \times [-4.5,4.5]$ (note that we did not generated trajectories of length $\geq 2$ in this case).
As validation data, we computed the values of $\bm{x}$ and $\dot{\bm{x}}$ analogously on the points of a $15 \times 15$ even grid in area $[-2,-2] \times [-4,4]$ (again not trajectories).
On the other hand, as test data, we generated $20$ trajectories of length $400$ with $\Delta t = 0.05$.

\paragraph{Model and Hyperparameter}
We selected the architectures of the model components by validation data loss.
The search range for each component was as follows:
(i) The search range for $\bm\psi$ in $\bm\phi$ (see \eqref{_eq:anode1} and \eqref{_eq:anode2}) was: 32, 64, 128, 32-32, \textcolor{red}{64-64}, and 128-128.
(ii) The search range for base dynamics $\bm{h}$ was: 32, 64, 128, \textcolor{red}{32-32}, 64-64, and 128-128.
(iii) The search range for the architecture of $q$ was: 32, 64, \textcolor{red}{128}, 256, 16-16, 32-32, 64-64, and 128-128.
Note that we did not perform the most intensive grid search.
Instead, we set (i) 32-32, (ii) 64-64, and (iii) 32-32 initially, and conducted a coordinate-descent-like search, that is, we updated (i)--(iii) in turn following validation loss values, until no further improvement was observed.

\paragraph{Optimization}
We used the Adam optimizer with learning rate $0.001$ and full-batch updates.
We set the decay rate parameter to be $0.0001$.
We adopted the early stopping strategy with validation loss.

\paragraph{Computational Speed}
For a full-batch update of gradient descent (i.e., a single epoch), the proposed method required about $40$ [ms] in average, whereas the vanilla method required about $15$ [ms]. 


\subsection{Application: fluid flow prediction}

\paragraph{Data}
We generated the data following the two-dimensional Navier--Stokes equations using a solver of the immersed boundary projection method \citep{Taira07,Colonius08}.
We followed the configuration specified in \citep[][Chapter 2]{dmdbook}.
The original generated trajectory starts at an unstable equilibrium and almost reaches to the limit cycle.
We retrieved 1,800 snapshots of the original trajectory with $\Delta t=0.2$ (note that the simulation was run with $\Delta t=0.02$, and we retrieved every 10 snapshots).
We used the first 600 snapshots as training data, which contained the observation from just after the unstable equilibrium to just before the limit cycle.
We used the next 200 snapshots as validation data, which contained the last transition phase to the limit cycle.
We used the last 1,000 snapshots as training data, which contained only the limit cycle behavior.
Whereas the simulation outputs the velocity computed on the grid points of $200 \times 450$, we used the values of vorticity computed from the velocity, hence a snapshot of the data was $199 \times 449 = 89351$ dimensional.
We reduced the dimensionality to $26$ using principal component analysis with only $0.1$\% energy loss.
After the dimensionality reduction, we normalized the data so that the largest absolute value becomes $1$.
We then added iid noise following Gaussian $\mathcal{N}(0,0.005^2)$.

\paragraph{Model and Hyperparameter}
We selected the architectures of the model components by validation data loss.
The search range for each component was as follows:
(i) The search range for $\bm\psi$ in $\bm\phi$ (see \eqref{_eq:anode1} and \eqref{_eq:anode2}) was: 16, \textcolor{red}{32}, 64, 128, 8-8, 16-16, and 32-32.
(ii) The search range for base dynamics $\bm{h}$ was: 256, 512, 1024, 2048, 32-32, 64-64, \textcolor{red}{128-128}, 256-256, and 512-512.
(iii) The search range for the architecture of $q$ was: 16, 32, 64, \textcolor{red}{128}, 256, 512, 1024, 16-16, 32-32, 64-64, 128-128, 256-256, and 512-512.
Note that we did not perform the most intensive grid search.
Instead, we set (i) 32, (ii) 128-128, and (iii) 32 initially, and conducted a coordinate-descent-like search, that is, we updated (i)--(iii) in turn following validation loss values, until no further improvement was observed.

\paragraph{Optimization}
We used the Adam optimizer with learning rate $0.0001$ and full-batch updates.
We set the decay rate parameter to be $0.0001$.
We adopted the early stopping strategy with validation loss.

\paragraph{Computational Speed}
For a full-batch update of gradient descent (i.e., a single epoch), the proposed method required about $100$ [ms], whereas the vanilla method required about $30$ [ms].

\end{document}